\documentclass[10pt,twocolumn,letterpaper]{article}
\PassOptionsToPackage{table}{xcolor}
\usepackage[pagenumbers]{cvpr}
\usepackage[utf8]{inputenc}
\usepackage[T1]{fontenc}
\usepackage{amsmath,amssymb,amsfonts}
\usepackage{booktabs,multirow,makecell}
\usepackage{nicefrac,microtype,verbatim,float,graphicx,wrapfig}
\usepackage{algorithm,algpseudocode}
\makeatletter
\AtBeginDocument{\providecommand{\theHALG@line}{}\renewcommand{\theHALG@line}{\thealgorithm.\arabic{ALG@line}}}
\makeatother
\graphicspath{{figure/}}

\definecolor{cvprblue}{rgb}{0.21,0.49,0.74}
\usepackage[pagebackref,breaklinks,colorlinks,allcolors=cvprblue]{hyperref}
\def\paperID{*****}
\def\confName{CVPR}
\def\confYear{2026}
\title{Sparse Attention for Video Generation Acceleration via Growing Sparsity and Reduced Search}
\author{
\textbf{Yun Dai}$^{1}$\quad
\textbf{Yuyun Chen}$^{1}$\quad
\textbf{Jiarui Wen}$^{1}$\quad
\textbf{Huiping Zhuang}$^{1}$\quad
\textbf{Cen Chen}$^{1}$\quad
\textbf{Ziqian Zeng}$^{1}$\thanks{Corresponding author.} \\
$^{1}$South China University of Technology \\
\texttt{zqzeng@scut.edu.cn}
}
\begin{document}
\maketitle
\begin{abstract}
Diffusion Transformers (DiTs) dominate video generation but suffer from high inference cost due to the quadratic complexity of self-attention.
Sparse attention mitigates this by skipping unnecessary computation in the attention matrix, but existing methods apply a fixed sparsity ratio throughout denoising and struggle to accurately determine which computations to skip.
To address these limitations, we analyze attention dynamics through attention entropy and mean attention distance, and find that attention becomes increasingly concentrated across denoising steps.
We further observe that blocks selected early continue to capture a large fraction of the total attention weight at later steps.
These findings suggest two acceleration opportunities: the sparsity ratio should grow as denoising progresses, and earlier search results can be reused to reduce the search space.
Based on these findings, we propose \textbf{Growing Sparsity and Reduced Search (GSRS)}, a training-free framework that retains more computation early and progressively reduces it later, while restricting subsequent searches to an early candidate set to reduce search overhead.
GSRS adopts \textit{Top-p} selection for adaptive sparsity control, supported by a precise evaluation scheme that computes exact attention weights within the FlashAttention computation at negligible overhead.
Experiments on HunyuanVideo and Wan 2.1 validate that GSRS achieves 1.55$\times$--2.21$\times$ speedup while maintaining generation quality comparable to full attention.
\end{abstract}

\section{Introduction}
\label{sec:intro}
Diffusion Transformers (DiTs) \cite{peebles2023scalable} represent the dominant architecture in the field of video generation \cite{chen2024far,ho2022video,fei2024dysen,lin2024open}, with models such as Sora \cite{liu2024sora}, HunyuanVideo \cite{kong2024hunyuanvideo}, and Wan \cite{wan2025wan} achieving state-of-the-art performance in high-resolution and long-duration synthesis.
Despite their success, the 3D full attention \cite{vaswani2017attention} has emerged as the primary computational bottleneck within this framework \cite{xiao2023combining}.
The 3D full attention converts video and text tokens into a single unified token sequence and apply self-attention across them, which causes inference costs to rise sharply as video resolution and temporal length increase.
For instance, generating a 125-frame 720p video using HunyuanVideo on a single NVIDIA A800 GPU requires processing more than 115K tokens and takes approximately 38 minutes, with self-attention operations accounting for roughly 80\% of the total latency.

Recent studies have shown that attention computation in DiTs contains substantial redundancy \cite{child2019generating,h2o,efficient,quest}, with the majority of token interactions making a negligible contribution to the final output.
Sparse attention methods \cite{guo2024block,zaheer2020big} exploit this redundancy by partitioning the attention matrix into fixed-size blocks and selecting a subset of blocks to compute.
Existing sparse attention strategies generally fall into two categories.
\textbf{Static methods} \cite{xi2025sparse,zhang2025fast,liuclear,li2025compact,li2025radial} restrict block interactions to fixed windows, and \textbf{dynamic methods} \cite{jiang2024minference,xu2025xattention,yangsparse,liu2021transformer,zhangtraining,tan2025dsv,pu2024efficient} identify critical blocks at runtime.
Critical blocks are blocks that actually contribute to the final output.
These approaches achieve notable speedups, yet they share two limitations.
First, a fixed sparsity ratio or selection threshold may allocate computation inefficiently across denoising steps.
Second, accurately and efficiently identifying critical blocks remains challenging.
Static methods rely on predefined blocks that may misalign with the ground truth.
Evaluating block importance on the fly is expensive for dynamic methods, which may reduce identification accuracy to improve evaluation efficiency.

A detailed analysis of how attention behavior changes during the denoising process can guide the design of a better sparse attention strategy.
To characterize this change, we measure how concentrated each query's attention is via \textit{attention entropy} \cite{araabi2024entropy} and how far each query attends to via \textit{mean attention distance} \cite{li2025attention}.
Lower entropy indicates that attention is focused on fewer tokens, while shorter distance means attention is more localized on nearby tokens.
We observe that both metrics generally decrease across denoising steps, meaning that attention focuses on fewer and closer tokens as denoising progresses.
This observation suggests \textbf{growing sparsity}.
Retaining more computation early and progressively reducing it later can improve the quality-efficiency trade-off.
Meanwhile, we find that a candidate set of blocks selected early in denoising continues to capture a large fraction of the total attention weight at later steps.
This finding suggests \textbf{reduced search}.
Subsequent searches can focus on the candidate set established at an early denoising step rather than repeatedly evaluating all blocks, reducing search overhead while maintaining high coverage of critical blocks.

Based on these insights, we propose \textbf{Growing Sparsity and Reduced Search (GSRS)}, a training-free framework that translates the above findings into two concrete acceleration strategies.
To realize \textbf{growing sparsity}, GSRS progressively lowers selection thresholds over denoising steps, retaining more computation early and allowing greater sparsity later.
To realize \textbf{reduced search}, GSRS establishes a candidate set of blocks at an early denoising step and restricts subsequent searches to this set, avoiding repeated evaluation of all blocks.
This helps reduce search overhead while maintaining high coverage of critical blocks.
Both mechanisms rely on selecting critical blocks at each search step.
GSRS adopts cumulative attention probability (\textit{Top-p}) \cite{holtzmancurious,twilight,tactic} to adaptively adjust the number of retained blocks based on the actual attention concentration, which requires precise measurement of block importance to correctly select critical blocks.
To support reliable \textit{Top-p} selection, we design a precise block importance evaluation scheme fused into the FlashAttention \cite{dao2022flashattention} computation at negligible overhead.
Together, these strategies allocate computation according to the changing attention concentration and reduce the cost of identifying critical blocks, improving the quality-efficiency trade-off.

Our contributions are summarized as follows:
\begin{itemize}
    \item We systematically analyze DiT attention using attention entropy, mean attention distance, and how much attention weight early-selected blocks retain at later denoising steps, providing empirical motivation for growing sparsity and reduced search.
    
    \item We propose \textbf{GSRS}, a training-free sparse attention framework that combines growing sparsity and reduced search.
    To support reliable \textit{Top-p} selection in GSRS, we design a precise block importance evaluation scheme fused into FlashAttention at negligible overhead.
    
    \item Extensive experiments on HunyuanVideo and Wan2.1 demonstrate that GSRS achieves up to 2.21$\times$ speedup while maintaining generation quality comparable to full attention, and delivers a superior quality-efficiency trade-off compared to compared sparse attention methods.
\end{itemize}

\section{Related Works}
\textbf{Sparse Attention in LLMs.}
Sparse attention has been extensively studied in large language models to address the quadratic complexity of self-attention over long sequences.
StreamingLLM \cite{xiaoefficient} introduces a sliding window with attention sinks for efficient streaming inference.
MInference \cite{jiang2024minference} identifies dynamic sparse patterns and performs online approximate search to select critical tokens.
MoBA \cite{lu2025moba} and NSA \cite{yuan2025native} further explore block-level importance estimation and hardware-aligned sparse designs for efficient long-context understanding.
While these methods provide valuable insights for sparse attention research, they are designed for single-forward inference and do not account for the multi-step iterative denoising process in DiTs, limiting their direct applicability to video generation.

\textbf{Sparse Attention in Video DiTs.}
Static methods \cite{xi2025sparse,liuclear,zhang2025fast} predefine spatial-temporal sparse patterns based on local windows or head-type classification, offering simplicity but limited adaptability to diverse attention behaviors.
Dynamic methods \cite{vorta,paroattention,zhangtraining,xia2025training,zhang2025spargeattention,xu2025xattention} perform runtime block selection through various importance estimation strategies, achieving better adaptability but struggling to balance identification accuracy with search overhead.
Despite their diverse strategies, these methods do not exploit how attention behavior changes across denoising steps.

\textbf{Other Acceleration Techniques for DiTs.}
Beyond sparse attention, several orthogonal techniques accelerate DiT inference.
Caching-based methods \cite{lvfastercache,ma2025magcache,chu2025omnicache,liu2025timestep,yang2023adacache,zhaoreal} reduce redundant computation by reusing intermediate activations across denoising steps.
Low-bit attention quantization methods \cite{stepbaq,gptq,duquant,awq,efficient,zhangsageattention} accelerate inference by reducing the numerical precision of attention computation.
Step distillation methods \cite{blade,imagen,dollar,zheng2025large,transition,zhang2025accvideo} reduce the number of sampling steps required.
Bottleneck Sampling \cite{tian2025training} employs variable resolution across different sampling stages to reduce token count during intermediate phases.
These approaches are complementary to sparse attention and can be combined with our method for further speedup.

\begin{figure*}[t]
    \centering
    \includegraphics[width=1.0\textwidth]{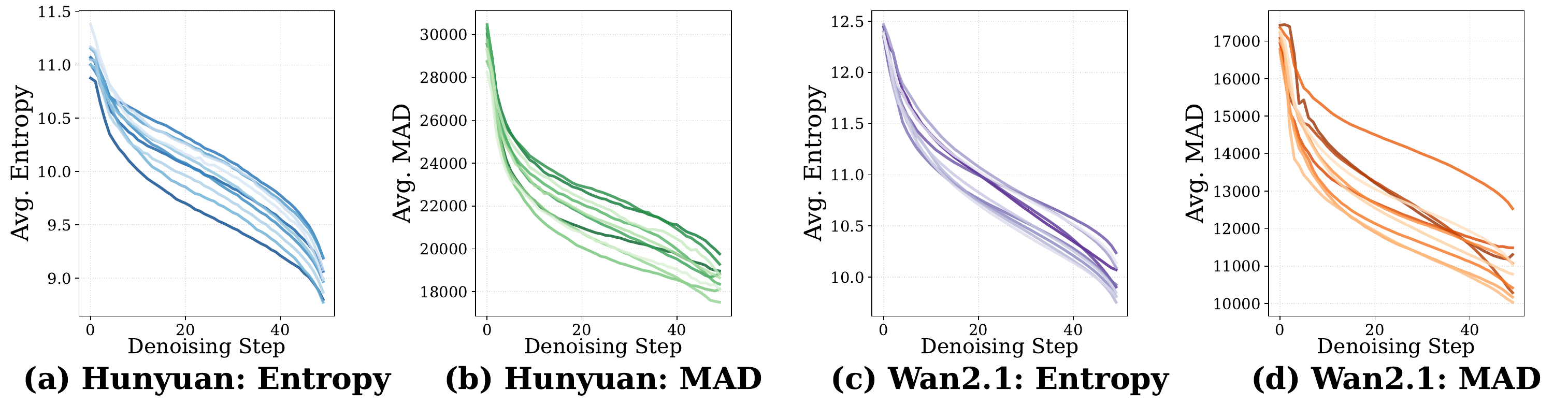}
    \caption{\textbf{Quantitative analysis of attention dynamics in DiT models.}
    The subfigures (a–d) displays the variation of attention entropy and mean attention distance (MAD) across 10 diverse prompts, where curves of different shades represent individual samples.
    Tokens are reordered via Hilbert curves to preserve spatial-temporal adjacency before computing positional distances (Appendix~\ref{sec:app_hilbert}).
    }
    \label{fig:attention_dynamics}
\end{figure*}

\label{gen_inst}
\section{Preliminary}
\label{headings}
\subsection{Attention and FlashAttention}
In the self-attention mechanism \cite{vaswani2017attention}, input tokens are projected into query, key, and value matrices $\mathbf{Q}, \mathbf{K}, \mathbf{V} \in \mathbb{R}^{L \times d}$, where $L$ is the sequence length and $d$ is the head dimension.
The attention output is computed as:
\begin{equation}
    \mathbf{W} = \mathrm{softmax}\!\left(\frac{\mathbf{Q}
    \mathbf{K}^\top}{\sqrt{d}}\right), \quad \mathbf{O} =
    \mathbf{W}\mathbf{V},
\end{equation}
which requires materializing the full $L\times L$ attention matrix, incurring ${O}({L^2})$ time and memory cost.
FlashAttention \cite{dao2022flashattention,daoflashattention,shah2024flashattention,zadouri2026flashattention} addresses this by partitioning $\mathbf{Q}, \mathbf{K}, \mathbf{V}$ into blocks and computing attention incrementally without materializing the full matrix.
During this process, it maintains a per-row Log-Sum-Exp (LSE) \cite{bahdanau2014neural} value as the global normalization factor, which we later leverage for precise block importance evaluation (§\ref{sec:precise}).
A detailed description of FlashAttention's computation is provided in Appendix~\ref{sec:app_flashattention}.

\subsection{Block Sparse Attention}
Block sparse attention reduces computation by skipping irrelevant blocks during the FlashAttention loop.
Given a \textbf{sparse mask} $\mathbf{M} \in \{0, 1\}^{L \times L}$, the sparse attention output is computed as:
\begin{equation}
    \mathbf{W} = \mathrm{softmax}\!\left(\frac{\mathbf{Q}\mathbf{K}^\top}{\sqrt{d}} \odot \mathbf{M}\right), \quad \mathbf{O} = \mathbf{W}\mathbf{V},
\end{equation}
where $\odot$ denotes element-wise masking that excludes unselected blocks from the softmax computation.
In practice, these blocks are simply skipped during the FlashAttention tiling loop, avoiding both computation and memory access.

We measure \textit{attention recall} as the fraction of total attention weight captured by the selected blocks:
\begin{equation}
    \mathrm{Recall}(\mathbf{A}, \mathbf{M})
    = \frac{\sum_{i,j} A_{ij} M_{ij}}{\sum_{i,j} A_{ij}},
\end{equation}
where $\mathbf{A} = \mathrm{softmax}(\mathbf{Q}\mathbf{K}^{\top}/\sqrt{d})$ denotes the full attention weights.
Higher recall indicates that the selected blocks capture more of the total attention weight.

\begin{figure}[H]
    \centering
    \includegraphics[width=\linewidth]{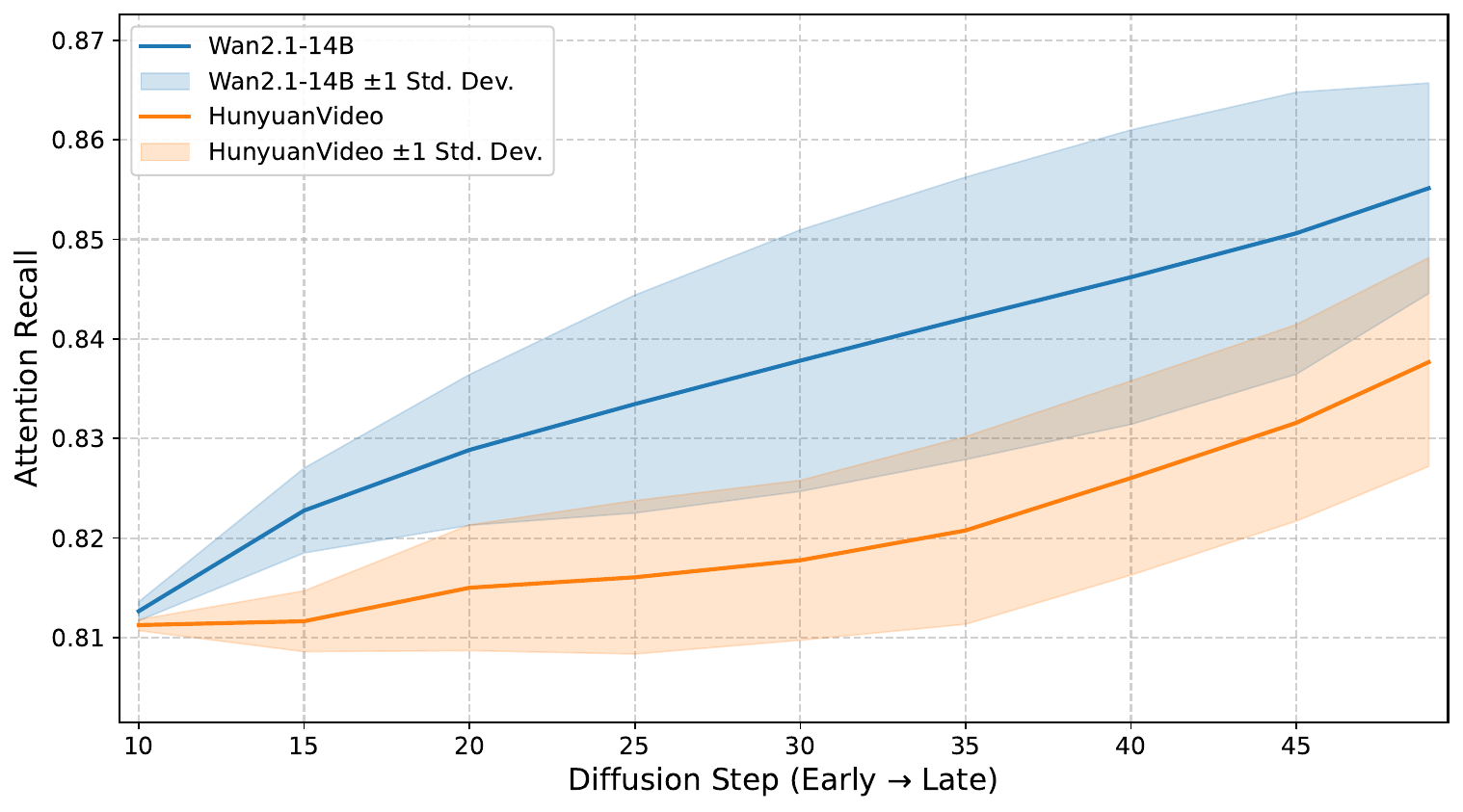}
    \caption{\textbf{Attention recall under a fixed sparse mask.}
    A sparse mask is constructed at an early denoising step and held fixed at subsequent steps.
    The mask maintains high attention recall throughout subsequent steps on both HunyuanVideo and Wan2.1.
    }
    \label{fig:attention_recall}
    \vspace{-10pt}
\end{figure}

\section{Attention Dynamics Analysis}
\label{sec:analysis}
Before presenting our method, we investigate how attention behavior changes during the denoising process.
To characterize these dynamics, we introduce two metrics based on the post-softmax attention weight $w_{ij}$ from query $i$ to key $j$.
Attention entropy $H(i)$ and mean attention distance $D(i)$ are defined respectively as:
\begin{equation}
    H(i) = -\sum_j w_{ij} \log w_{ij}, \quad D(i) = \sum_j w_{ij} \cdot |i - j|,
\end{equation}
$H(i)$ measures how concentrated the attention distribution is.
Lower entropy means attention is focused on fewer tokens.
$D(i)$ measures the weighted average distance between the query $i$ and the keys $j$ it attends to, where $|i - j|$ denotes the distance between positions $i$ and $j$ in the token sequence.
Smaller mean attention distance means attention is focused on nearby tokens.

We conduct systematic analysis on HunyuanVideo and Wan2.1 across multiple prompts, yielding the following key observations.

\begin{figure*}[t]
    \centering
    \includegraphics[width=1.0\textwidth]{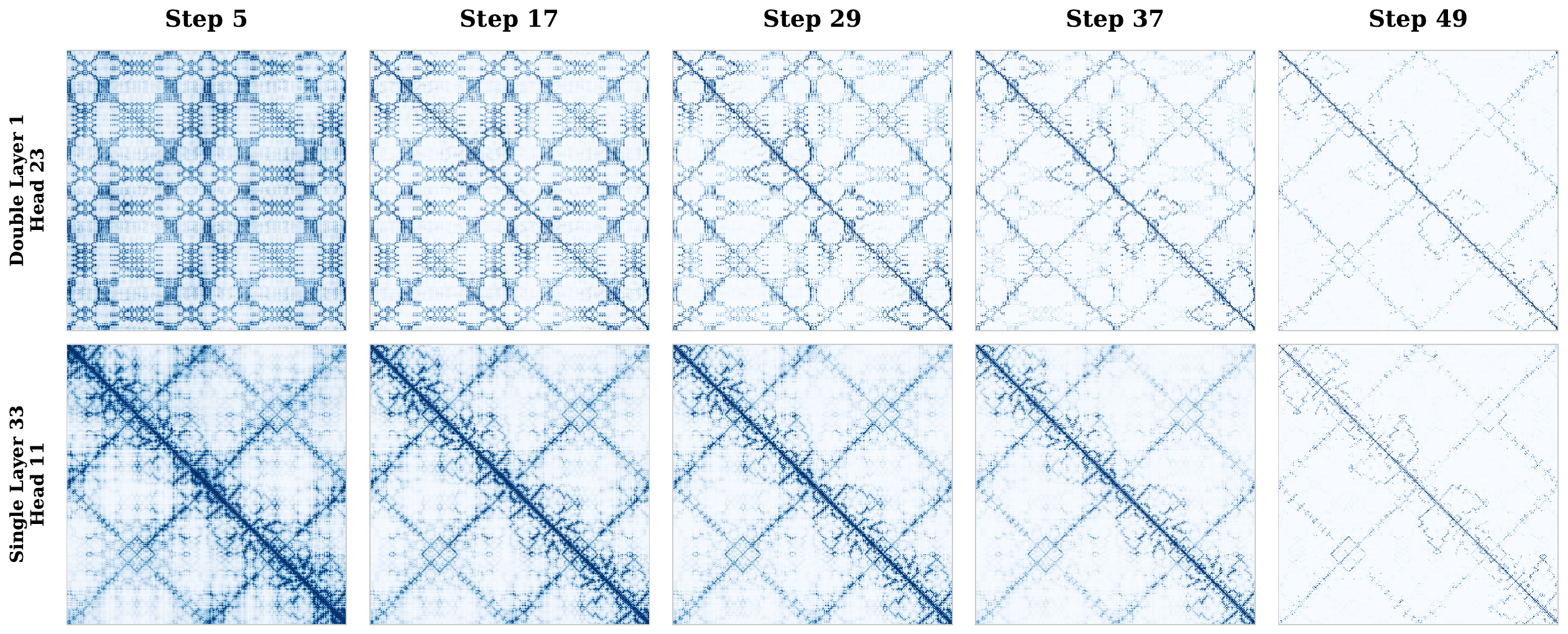} 
    \caption{\textbf{Block-level attention patterns across denoising steps.}
    Each cell represents the attention weight between a query block and a key block.
    Two representative heads at steps 5, 17, 29, 37, and 49 on HunyuanVideo.
    Tokens are reordered via Hilbert curves \cite{cerveny2025gilbert} to preserve spatial-temporal adjacency.
    As attention becomes more concentrated, blocks with high attention weights early in denoising continue to capture substantial attention weight, while those with low weights tend to retain low weights at later steps.
    }
    \label{fig:attn_evolution}
\end{figure*}

\paragraph{Observation 1: Attention entropy and distance generally decrease across denoising steps.}\label{obs:entropy}
As shown in Figure~\ref{fig:attention_dynamics}(a-d), both attention entropy and mean attention mean attention distance generally decrease across denoising steps on HunyuanVideo and Wan2.1.
This indicates that attention focuses on fewer and closer tokens as denoising progresses.
This trend remains consistent across different model architectures and independent of specific inputs, indicating an inherent property of the denoising process.
\textbf{This motivates us to allocate more computation to early denoising steps and progressively increase sparsity later to improve the quality-efficiency trade-off.}

\paragraph{Observation 2: An early candidate set of blocks maintains high attention recall at later steps.}\label{obs:stability}
Figure~\ref{fig:attn_evolution} illustrates that blocks with high attention weights early in denoising continue to capture substantial attention weight at later steps.
Blocks with low attention weights early in denoising also tend to retain low weights at later steps.
To quantify this observation, we select a candidate set of blocks at an early denoising step and construct a sparse mask from this set.
We then keep this mask fixed and measure its attention recall at subsequent denoising steps.
As shown in Figure~\ref{fig:attention_recall}, this candidate set continues to capture a large fraction of the total attention weight on both HunyuanVideo and Wan2.1.
\textbf{This motivates us to restrict subsequent searches to the candidate set established early in denoising, reducing search overhead while maintaining high coverage of critical blocks.}

\section{Methodology}
\begin{figure*}[t]
    \centering
    \includegraphics[width=1.0\textwidth]{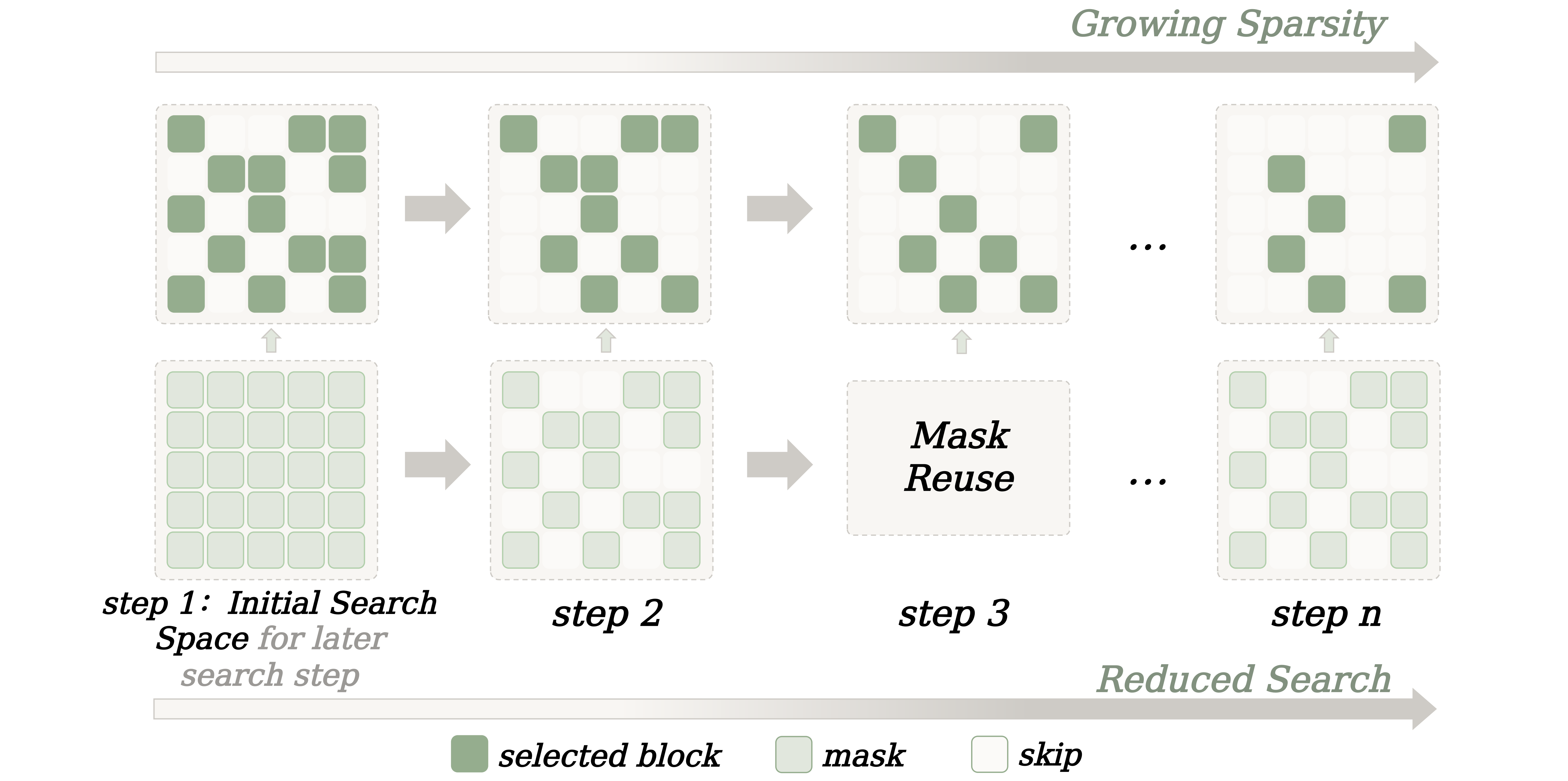} 
    \small 
    \caption{\textbf{Overview of GSRS.}
    The top row illustrates \textbf{growing sparsity}.
    As denoising progresses, progressively tighter \textit{Top-p} thresholds retain fewer blocks, yielding increasingly sparse masks.
    The bottom row illustrates \textbf{reduced search}.
    The first search step evaluates all blocks to establish a baseline sparse mask.
    Subsequent search steps only evaluate blocks within this baseline mask, while non-search steps directly reuse the most recent mask without re-evaluation.}
    \label{fig:overview}
\end{figure*}

Guided by the analysis in Section~\ref{sec:analysis}, we propose \textbf{GSRS}, a training-free framework that applies increasing sparsity across denoising steps while reducing the search space for block selection.
We visualize the workflow of GSRS in Figure~\ref{fig:overview}.
At the first search step, GSRS evaluates all blocks and establishes a baseline sparse mask.
All subsequent searches are restricted to this baseline mask with progressively tighter \textit{Top-p} thresholds (§\ref{sec:progressive}).
Between search steps, the most recent mask is directly reused.
Central to this framework is \textit{Top-p} block selection for adaptive sparsity control (§\ref{sec:top_p}), which requires precise measurement of each block's importance for reliable selection.
To support reliable \textit{Top-p} selection, we propose a precise block importance evaluation scheme fused into the FlashAttention computation at negligible overhead (§\ref{sec:precise}).

\subsection{Growing Sparsity and Reduced Search}
\label{sec:progressive}
The search strategy of GSRS comprises two complementary mechanisms: growing sparsity that increases the sparsity ratio across denoising steps to retain more computation early and reduce it later, and reduced search that limits the search space to an early candidate set while preserving dynamic block selection within this set.

\paragraph{Growing Sparsity.}
As revealed by Observation \hyperref[obs:entropy]{1}, attention entropy and mean attention distance generally decrease across denoising steps.
This motivates allocating more computation to early steps and increasing sparsity later.
GSRS employs cumulative attention probability (\textit{Top-p}) for block selection.
Blocks are sorted by their importance in descending order and retained until their cumulative importance exceeds a threshold $p$.
Inspired by the observation, GSRS applies a decreasing threshold $\{p_t\}$ across search steps, starting from a high $p_{\text{start}}$ and gradually lowering to $p_{\text{end}}$.
Early steps retain more blocks because the model's attention is still spread widely, while later steps safely discard more blocks as attention concentrates on fewer tokens.
As validated in Table~\ref{tab:p_schedule}, this decreasing threshold schedule consistently outperforms both fixed and shuffled $p$ configurations under comparable computational budgets.

\paragraph{Reduced Search.}
Observation \hyperref[obs:stability]{2} shows that a candidate set of blocks selected early in denoising continues to capture a large fraction of the total attention weight at later steps.
GSRS uses this candidate set to limit subsequent searches while maintaining high coverage of critical blocks.
Given a predefined set of search steps $\mathcal{T} = \{t_1, t_2, \ldots, t_n\}$, GSRS begins with full attention warmup for the first $N_w$ steps.
At step $t_1$, a full-space evaluation establishes a baseline sparse mask $\mathcal{S}_1$.
The blocks selected by $\mathcal{S}_1$ form a fixed candidate set for subsequent searches.
At each subsequent search step $t_k$ ($k \geq 2$), GSRS re-evaluates all blocks within $\mathcal{S}_1$ instead of restricting selection to the previously selected mask $\mathcal{S}_{k-1}$.
It then applies threshold $p_k$ to obtain an updated sparse mask $\mathcal{S}_k$, allowing previously omitted blocks within $\mathcal{S}_1$ to be selected again.
Between search steps, the most recently updated sparse mask is directly reused without re-evaluation.
This design preserves the ability to dynamically select blocks while reducing search overhead compared with evaluating all blocks at every search step, as validated in Table~\ref{tab:cached_search}.

\subsection{Adaptive Block Selection via Top-p.}
\label{sec:top_p}
Prior works \cite{xia2025training,li2025compact,xi2025sparse} have shown that sparse patterns varies substantially across layers and heads.
A fixed \textit{Top-k} strategy assigns the same number of blocks to every head, regardless of its actual sparse pattern.
We also observe this variation in our experiments and provide detailed analysis in Appendix~\ref{sec:app_layers_analysis} and \ref{sec:app_head_analysis}.
To address this, GSRS employs cumulative attention probability (\textit{Top-p}) for block selection.
Blocks are sorted by their importance in descending order and retained until their cumulative importance exceeds a threshold $p$ (Algorithm~\ref{alg:top_p}).
For heads where a few blocks carry most of the attention weight, \textit{Top-p} retains only those few blocks.
For heads where attention is distributed more evenly, more blocks are needed to reach the same threshold $p$.
A single threshold $p$ thus achieves adaptive sparsity across layers and heads without any manual tuning.
The advantage of \textit{Top-p} over fixed \textit{Top-k} is validated in Section~\ref{sec:ablation}.

\subsection{Precise Block Importance Evaluation}
\label{sec:precise}
The \textit{Top-p} selection described above requires precise measurement of each block's importance to determine which blocks to retain and how many are needed to reach the cumulative probability threshold.
In this section, we explain why obtaining this measurement within FlashAttention is challenging, and present our two-phase evaluation scheme.

\paragraph{Block Importance Definition.}
Given a query block $u$ and a key block $v$, each containing $B$ tokens, we define the block importance $W_{uv}$ as the average attention weight that query block $u$ assigns to key block $v$:
\begin{equation}
\label{eq:block_importance}
\begin{aligned}
    W_{uv}
    &= \frac{1}{B}\sum_{i\in u}\sum_{j\in v}w_{ij} \\
    &= \frac{1}{B}\sum_{i\in u}\sum_{j\in v}\exp(z_{ij}-L_i).
\end{aligned}
\end{equation}
Here, $z_{ij}=\mathbf{q}_i^\top\mathbf{k}_j/\sqrt{d}$ is the scaled attention score between query $i$ and key $j$, and $w_{ij}$ denotes the corresponding post-softmax attention weight.
The row-wise normalization factor $L_i=\mathrm{LSE}(\mathbf{z}_i)$ is computed via Eq.~\ref{eq:lse}, where $\mathbf{z}_i=[z_{i1},z_{i2},\ldots,z_{iL}]$ contains the attention scores between query $i$ and all key tokens.
For each query block $u$, the block importance $W_{uv}$ sums to 1 over all key blocks $v$.
This property is essential for \textit{Top-p} selection, which relies on accumulating block importance until the sum reaches the threshold $p$.

\paragraph{The Challenge of Computing Precise Block Importance.}
Computing $W_{uv}$ requires the global normalization factor $L_i$, which depends on the scores between query $i$ and \textit{all} key tokens.
However, FlashAttention processes key blocks one at a time and only produces the final $L_i$ after the last key block has been processed (see Appendix~\ref{par:lse_availability}).
This means that during the computation, we cannot yet compute exact attention weights for any individual block.
Prior work \cite{xia2025training} addresses this by reusing the $L_i$ cached from a previous denoising step, denoted $L_i^{\text{old}}$, to compute an approximate block importance:
\begin{equation}
    \hat{W}_{uv} = \frac{1}{B}\sum_{i\in u}\sum_{j\in v}\exp(z_{ij} - L_i^{\text{old}}),
\end{equation}
However, as the attention distribution changes across denoising steps, the true normalization factor $L_i^{\text{new}}$ drifts away from the cached $L_i^{\text{old}}$.
This drift scales each row's block importance by a factor of $\exp(L_i^{\text{new}} - L_i^{\text{old}})$, causing them to no longer sum to 1.
Since \textit{Top-p} selects blocks until their importance sums exceed a threshold $p$, inaccurate sums directly affect the selection.
When the sums exceed 1, the threshold is reached too early, causing important blocks to be incorrectly discarded.

\paragraph{Two-Phase Block Importance Evaluation.}
GSRS computes block importance in two phases without relying on cached LSE values.
During FlashAttention computation, we record the log-sum-exp of attention scores for each query row within each key block.
Once the current row-wise LSE becomes available, we normalize these stored statistics and average the normalized weights over query rows to obtain block importance.

\textit{Phase 1: Fused block log-sum-exp computation.}
During FlashAttention computation, we compute and store the log-sum-exp of attention scores $G_{iv}$ within each key block:
\begin{equation}
    G_{iv} = m_{iv} + \log \sum_{j \in v} \exp(z_{ij} - m_{iv}),
\end{equation}
where $m_{iv}=\max_{j\in v}z_{ij}$ is the maximum score within key block $v$ for query $i$.
Computing $\exp(z_{ij}-m_{iv})$ keeps the exponent non-positive, preventing numerical overflow.
We store only one scalar $G_{iv}$ per query row and key block, avoiding storage of the full $L\times L$ attention matrix.

\textit{Phase 2: Block importance recovery.}
After FlashAttention finishes and the current row-wise LSE $L_i$ is available, we recover block importance as
\begin{equation}
\begin{aligned}
    W_{uv}
    &= \frac{1}{B}\sum_{i\in u}\exp(G_{iv}-L_i) \\
    &= \frac{1}{B}\sum_{i\in u}\sum_{j\in v}\exp(z_{ij}-L_i) \\
    &= \frac{1}{B}\sum_{i\in u}\sum_{j\in v}w_{ij}.
\end{aligned}
\end{equation}
This computation recovers the exact block importance defined in Eq.~\eqref{eq:block_importance} without using cached LSE.
A detailed derivation is provided in Appendix~\ref{sec:app_block_importance_proof}.

Phase 1 is fused into FlashAttention, while Phase 2 requires only lightweight post-processing of the stored statistics.
Together, they require no additional pass over the data and introduces negligible overhead in both latency (Figure~\ref{fig:kernel_latency}) and memory (Table~\ref{tab:memory}).
The complete procedure is detailed in Appendix~\ref{sec:app_fused_eval}.

\paragraph{Differences from AdaSpa and LiteAttention.}
AdaSpa \cite{xia2025training} uses a fixed sparsity target and evaluates all blocks at scheduled search steps using cached LSE values.
GSRS instead progressively lowers Top-p thresholds and restricts subsequent searches to $\mathcal{S}_1$, reducing both attention computation and search overhead.
Moreover, GSRS computes block importance from block log-sum-exp statistics and the current row-wise LSE, avoiding errors caused by cached LSE values and supporting reliable \textit{Top-p}/\textit{Top-k} selection.
Once LiteAttention \cite{liteattention} skips a block, it excludes that block from both evaluation and computation for the remaining denoising steps.
GSRS re-evaluates all blocks within $\mathcal{S}_1$ at each search step, allowing previously omitted blocks to be selected again as attention changes.
See Appendix~\ref{sec:comparison} for more details.

\begin{table*}[t]
    \centering
    \renewcommand{\arraystretch}{1.4}
    \caption{\textbf{Quantitative evaluation of quality and efficiency.}
    We evaluate on Wan 2.1 and HunyuanVideo.
    \textbf{Bold} and \underline{underline} denote the best and second best results.}
    \label{tab:main_results}
    \resizebox{\textwidth}{!}{

    \begin{tabular}{ll!{\color{gray!40}\vrule width 0.4pt}cccccc!{\color{gray!40}\vrule width 0.4pt}ccc}
    \toprule
    \multirow{2.5}{*}{\textbf{Model}} & \multirow{2.5}{*}{\textbf{Method}} & \multicolumn{6}{c!{\color{gray!40}\vrule width 0.4pt}}{\textbf{Quality}} & \multicolumn{3}{c}{\textbf{Efficiency}} \\
    \cmidrule(lr){3-8} \cmidrule(lr){9-11}
     &                                            & PSNR $\uparrow$                  & SSIM $\uparrow$                  & LPIPS $\downarrow$   & SubConsist $\uparrow$              & ImageQual $\uparrow$               & VBench-Q $\uparrow$                    & Density $\downarrow$               & Latency $\downarrow$                & Speedup $\uparrow$                         \\
    \midrule

    \multirow{5}{*}{\textbf{Wan 2.1}}
     & \cellcolor{gray!9}Full Attention           & \cellcolor{gray!9}-              & \cellcolor{gray!9}-              & \cellcolor{gray!9}-- & \cellcolor{gray!9}98.26\%          & \cellcolor{gray!9}73.29\%          & \cellcolor{gray!9}83.75\%              & \cellcolor{gray!9}100\%            & \cellcolor{gray!9}3689s             & \cellcolor{gray!9}1.00$\times$             \\
     & SpargeAttn \cite{zhang2025spargeattention} & 20.55                            & 0.757                            & 0.295                   & 94.88\%                            & \underline{72.80\%}                & 82.21\%                                & 42.00\%                            & 2732s                               & 1.35$\times$                               \\
     & SVG \cite{xi2025sparse}                    & 20.87                            & 0.786                            & 0.277                   & 97.74\%                            & 71.99\%                            & \underline{83.69\%}                    & 40.14\%                            & 2675s                               & 1.38$\times$                               \\
     & SVG 2 \cite{yangsparse}                    & 22.83                            & 0.847                            & 0.214                   & 97.80\%                            & 72.23\%                            & 83.19\%                                & \underline{32.08\%}                & \underline{2455s}                               & \underline{1.50$\times$}                               \\
     & SVOO \cite{svoo}                           & \underline{23.34}                & \underline{0.865}                & \underline{0.195}                   & \underline{97.89\%}                & 72.42\%                            & 83.33\%                                & 32.82\%                            & 2610s                                 & 1.41$\times$                                 \\

     & \cellcolor{blue!8}\textbf{GSRS}            & \cellcolor{blue!8}\textbf{23.79} & \cellcolor{blue!8}\textbf{0.872} & \cellcolor{blue!8}\textbf{0.187} & \cellcolor{blue!8}\textbf{98.21\%} & \cellcolor{blue!8}\textbf{73.22\%} & \cellcolor{blue!8}\textbf{83.72\%}     & \cellcolor{blue!8}\textbf{30.63\%} & \cellcolor{blue!8}\textbf{2380s}    & \cellcolor{blue!8}\textbf{1.55$\times$}    \\
    \midrule

    \multirow{5}{*}{\textbf{Hunyuan}}
     & \cellcolor{gray!9}Full Attention           & \cellcolor{gray!9}-              & \cellcolor{gray!9}-              & \cellcolor{gray!9}-- & \cellcolor{gray!9}96.82\%          & \cellcolor{gray!9}70.71\%          & \cellcolor{gray!9}83.98\%              & \cellcolor{gray!9}100\%            & \cellcolor{gray!9}3057s             & \cellcolor{gray!9}1.00$\times$             \\
     & SpargeAttn \cite{zhang2025spargeattention} & 27.79                            & 0.907                            & 0.192                   & 96.55\%                            & \underline{70.65\%}                & \underline{83.85\%}                                & 33.00\%                            & 2014s                               & 1.52$\times$                               \\
     & SVG \cite{xi2025sparse}                    & 27.64                            & 0.895                            & 0.221                   & \underline{96.71\%}                & 68.98\%                            & 83.21\%                                & 31.65\%                            & 2066s                               & 1.48$\times$                               \\
     & SVG 2 \cite{yangsparse}                    & 29.27                            & 0.914                            & 0.213                   & 96.60\%                            & 66.50\%                            & 82.24\%                                & \underline{26.38\%}                & 1978s                               & \underline{1.55$\times$}                               \\
     & SVOO \cite{svoo}                           & \underline{30.31}                & \underline{0.931}                & \underline{0.172}                   & \textbf{96.88\%}                   & 69.19\%                            & 82.97\%                                & 27.12\%                            & \underline{1975s}                               & \underline{1.55$\times$}                               \\
     & \cellcolor{blue!8}\textbf{GSRS}            & \cellcolor{blue!8}\textbf{32.37} & \cellcolor{blue!8}\textbf{0.957} & \cellcolor{blue!8}\textbf{0.138} & \cellcolor{blue!8}96.66\%          & \cellcolor{blue!8}\textbf{70.79\%} & \cellcolor{blue!8} \textbf{84.01\%}    & \cellcolor{blue!8}\textbf{25.06\%} & \cellcolor{blue!8}\textbf{1825s}    & \cellcolor{blue!8}\textbf{1.68$\times$}    \\
    \bottomrule
    \end{tabular}
    }
\end{table*}

\section{Experiments}
\subsection{Experimental Setup}
\label{sec:exp_setup}

\paragraph{Models and Datasets.}
We evaluate GSRS on two open-sourced DiT models, Wan2.1-T2V-14B \cite{wan2025wan} and HunyuanVideo-T2V-13B \cite{kong2024hunyuanvideo}. 
Both generate 720p videos with 50 denoising steps. 
In latent space, Wan2.1 and HunyuanVideo process 21 and 32 frames respectively, with 3,600 tokens per frame.
For all experiments, we use the original dataset from VBench \cite{huang2023vbench} for both models.

\paragraph{Metrics.}
We use PSNR \cite{psnr}, SSIM \cite{ssim} and LPIPS \cite{lpips} to measure pixel-level similarity between videos generated with sparse attention and full attention. 
For video quality evaluation, we report Subject Consistency, Imaging Quality, and the overall quality score VBench-Q from VBench \cite{huang2023vbench}. 
To quantify the computational saving of sparse attention, we report density, defined as the ratio of actual computation performed to that of full attention. 
For efficiency, we report the end-to-end latency and the overall speedup.

\paragraph{Baselines.} 
We compare GSRS against both static and dynamic sparse attention methods. 
For static methods, we include SVG \cite{xi2025sparse}. 
For dynamic methods, we include SpargeAttention \cite{zhang2025spargeattention}, SVG 2 \cite{yangsparse} and SVOO \cite{svoo}. 
All baselines are evaluated under the same hardware configuration and input settings as GSRS. 

\paragraph{Implementation Details.}
We implement GSRS using TileLang \cite{tilelang2024} for fused attention computation and block importance evaluation at search steps, and the CUDA kernels from Block Sparse Attention \cite{guo2024block} for sparse attention computation at non-search steps.
All experiments are conducted on a single NVIDIA A800 GPU with CUDA 12.8.
Tokens are reordered using the Generalized Hilbert curve \cite{cerveny2025gilbert} with a block size of $B = 128$.
By default, we perform search at denoising steps $\mathcal{T} = \{10, 20, 30, 40\}$ with \textit{Top-p} thresholds $p = \{0.95, 0.9, 0.85, 0.8\}$.
For all experiments, the first 20\% of denoising steps are executed with full attention as warmup before applying sparse computation.
Experiments with different warmup lengths are provided in Appendix~\ref{sec:warmup_length}.

\begin{table*}[t]
\centering
\footnotesize
\setlength{\tabcolsep}{3pt}

\begin{minipage}[t]{0.49\textwidth}
    \centering
    \begin{minipage}[t][3.5\baselineskip][t]{\linewidth}
        \vspace{0pt}
        \caption{Effect of \textit{Top-p} schedule configurations on HunyuanVideo.}
        \label{tab:p_schedule}
    \end{minipage}\par
    \begin{tabular}{lcccc}
    \toprule
    $p$ schedule & PSNR$\uparrow$  & VB-Q$\uparrow$ & Latency$\downarrow$ \\
    \midrule
    \{0.95, 0.90, 0.85, 0.80\} & 31.96  & 84.99 & 1825 \\
    \{0.95, 0.80, 0.85, 0.90\} & 31.64  & 84.98 & 1850 \\
    \{0.93, 0.93, 0.93, 0.93\} & 31.20  & 84.90 & 1852 \\
    \{0.90, 0.90, 0.90, 0.90\} & 30.28  & 84.89 & 1774 \\
    \bottomrule
    \end{tabular}
\end{minipage}
\hfill
\begin{minipage}[t]{0.49\textwidth}
    \centering
    \begin{minipage}[t][3.5\baselineskip][t]{\linewidth}
        \vspace{0pt}
        \caption{Ablation on search strategies and block importance evaluation on HunyuanVideo.
        Den. represents Density.}
        \label{tab:cached_search}
    \end{minipage}\par
    \begin{tabular}{lcccc}
    \toprule
    Config & PSNR$\uparrow$ & VB-Q$\uparrow$ & Den.$\downarrow$ & Latency$\downarrow$ \\
    \midrule
    GSRS & 31.96 & 84.99 & 25.06 & 1825 \\
    w/ Cached LSE & 25.32 & 84.04 & 20.89 & 1730 \\
    w/ Previous-mask Search & 30.87 & 84.80 & 23.06 & 1757 \\
    w/o Reduced Search & 32.57 & 84.98 & 28.62 & 1990 \\
    \bottomrule
    \end{tabular}
\end{minipage}
\end{table*}

\begin{table*}[t]
\centering
\footnotesize
\setlength{\tabcolsep}{3pt}
\begin{minipage}[t]{0.49\textwidth}
    \centering
    \caption{Ablation on search steps $\mathcal{T}$ on HunyuanVideo.}
    \label{tab:search_step}
    \begin{tabular}{lcccc}
    \toprule
    $\mathcal{T}$ & PSNR$\uparrow$ & VB-Q$\uparrow$ & Latency$\downarrow$ \\
    \midrule
    \{10\} & 32.94 & 84.93 & 2076 \\
    \{10, 30\} & 32.59 & 84.88 & 1870 \\
    \{10, 20, 30\} & 31.89 & 84.88 & 1845 \\
    \{10, 30, 40\} & 32.59 & 84.89 & 1882 \\
    \{10, 20, 30, 40\} & 31.96 & 84.99 & 1825 \\
    \bottomrule
    \end{tabular}
\end{minipage}
\hfill
\begin{minipage}[t]{0.49\textwidth}
    \centering
    \caption{Comparison of \textit{Top-p} and \textit{Top-k} selection strategies on HunyuanVideo.}
    \label{tab:topk_vs_topp}
    \begin{tabular}{lcccc}
    \toprule
    Strategy & PSNR$\uparrow$ & VB-Q$\uparrow$ & Latency$\downarrow$ \\
    \midrule
    \textit{Top-p} (ours) & 31.96  & 84.99 & 1825 \\
    \textit{Top-k} (25\%) & 29.84  & 84.95 & 1813 \\
    \textit{Top-k} (50\%) & 32.67  & 85.03 & 2245 \\
    \bottomrule
    \end{tabular}
\end{minipage}
\end{table*}

\subsection{Main Results}
\label{sec:main_results}

\paragraph{Similarity.}
As shown in Table~\ref{tab:main_results}, GSRS achieves the highest PSNR and SSIM on both models, reaching an average PSNR of 32.75 on HunyuanVideo and 23.79 on Wan2.1, outperforming all baselines. 
The high similarity scores indicate that the sparse attention output of GSRS closely approximates full attention, preserving fine-grained details in the generated videos.  
Figure~\ref{fig:psnr_distribution} further shows that GSRS also achieves a more stable distribution, indicating more consistent performance across diverse prompts.

\paragraph{Quality.} 
GSRS maintains VBench-Q scores close to full attention and outperforms all evaluated sparse attention baselines on both models.
GSRS achieves the highest VBench-Q among the evaluated sparse attention methods on both models, with scores of 83.72\% on Wan2.1 and 84.01\% on HunyuanVideo, comparable to 83.75\% and 83.98\% for full attention. 
Full VBench-Q results are provided in Appendix~\ref{sec:vbench_result}.

\paragraph{Efficiency.}
GSRS achieves the lowest attention density and the highest speedup among the evaluated sparse attention methods on both models.
It reduces density to 30.63\% on Wan2.1 and 25.06\% on HunyuanVideo, achieving end-to-end inference speedups of \textbf{1.55$\times$} and \textbf{1.68$\times$}, respectively.
As shown in Appendix~\ref{sec:h800_speed}, GSRS also achieves the highest speedup among the evaluated methods on NVIDIA H800, reaching \textbf{1.78$\times$} on Wan2.1 and \textbf{2.21$\times$} on HunyuanVideo.

\subsection{Ablation Study}
\label{sec:ablation}
We conduct ablation studies on HunyuanVideo to validate each design choice of GSRS.
All ablations use the default configuration with search steps $\mathcal{T} = \{10, 20, 30, 40\}$ and cumulative probability thresholds $p = \{0.95, 0.9, 0.85, 0.8\}$.
Experiments are evaluated on a fixed subset of the VBench dataset.
VB-Q represent VBench Quality (VBench-Q).
Since PSNR measures pixel-level deviation while VBench-Q better reflects perceptual quality, we consider both metrics with emphasis on VBench-Q for quality assessment.

\paragraph{Decreasing $p$ Schedule.}
As shown in Table~\ref{tab:p_schedule}, the decreasing schedule achieves the highest PSNR and VBench-Q among the compared configurations.
The shuffled schedule lowers PSNR to 31.64 at similar latency, highlighting the benefit of decreasing thresholds over denoising steps.
Compared with the fixed schedule with $p=0.93$, the decreasing schedule improves PSNR from 31.20 to 31.96 while reducing latency from 1852s to 1825s.
These results support retaining more computation early and progressively reducing it later to improve the quality-efficiency trade-off.
Appendix~\ref{sec:top_p_configurations} provides results for additional \textit{Top-p} configurations.

\paragraph{Search Strategies.}
With previous-mask search, each search step is restricted to the mask $\mathcal{S}_{k-1}$ selected at the preceding search step, so previously excluded blocks cannot be selected again.
Without Reduced Search, each search step evaluates all blocks in the full attention matrix instead of restricting to the baseline mask $\mathcal{S}_1$.
Table~\ref{tab:cached_search} shows that removing Reduced Search increases density from 25.06\% to 28.62\% and latency from 1825s to 1990s while maintaining comparable VBench-Q.
As shown in Figure~\ref{fig:runtime_breakdown}, reduced search lowers the total search overhead by 13.75$\times$ on HunyuanVideo, confirming that restricting searches to the baseline mask is both safe and efficient. 
Previous-mask search reduces latency to 1757s but lowers PSNR and VBench-Q to 30.87 and 84.80, respectively, highlighting the benefit of preserving dynamic block selection within $\mathcal{S}_1$.

\paragraph{Precise Block Importance Evaluation.} 
We replace our precise block importance evaluation scheme with an estimator that uses cached LSE values from a previous denoising step.
As shown in Table~\ref{tab:cached_search}, this reduces PSNR from 31.96 to 25.32 and VBench-Q from 84.99 to 84.04.
Cached LSE values distort block importance and can alter both cumulative probability sums and block rankings, leading to inaccurate selection.
These results demonstrate the effectiveness of our precise block importance evaluation scheme in supporting reliable \textit{Top-p} selection.
Our scheme also enables accurate \textit{Top-k} block selection, as further evaluated in Appendix~\ref{sec:comparison}.

\paragraph{Top-p vs Top-k.} 
Table~\ref{tab:topk_vs_topp} compares \textit{Top-p} with \textit{Top-k} under matched density. 
\textit{Top-k} (25\%) matches the average density of \textit{Top-p} but only achieves a PSNR of 29.84, significantly lower than \textit{Top-p}'s 31.96 at similar latency. 
\textit{Top-k} (50\%) achieves slightly higher PSNR but incurs substantially higher latency of 1304s with only marginal improvement in VBench-Q. 
This confirms that adaptive \textit{Top-p} selection allocates computation more effectively than fixed budgets.

\paragraph{Search Steps.} 
Table~\ref{tab:search_step} shows that all configurations achieve high PSNR above 31.89 and stable VBench-Q scores, indicating that the generation quality is well preserved regardless of the number of search steps. 
Among them, the four-step configuration $\{10, 20, 30, 40\}$ achieves the highest VBench-Q score while maintaining low latency, resulting in the best overall trade-off between quality and efficiency.

\section{Conclusion}
\label{sec:conclusion}
In this paper, we propose GSRS, a training-free sparse attention framework featuring grow sparsity and reduced search to accelerate video DiT inference.
Combined with \textit{Top-p} and precise block importance evaluation, GSRS achieves up to 2.21$\times$ speedup while maintaining generation quality comparable to full attention.

\clearpage
{\small
\bibliographystyle{ieeenat_fullname}
\bibliography{reference}
}
\clearpage
\onecolumn
\raggedbottom
\begin{center}
\Large\bfseries\thetitle\\[0.5em]
\normalfont\Large Supplementary Material
\end{center}
\appendix
\section{Algorithm Details}
\label{sec:algorithm_details}

\subsection{FlashAttention Details}
\label{sec:app_flashattention}

\paragraph{Standard Softmax and Numerical Instability.}
Given a row of pre-softmax scores $\mathbf{z} = [z_1, z_2, \ldots, z_N]$, the standard softmax computes:
\begin{equation}
\text{softmax}(z_j) = \frac{\exp(z_j)}{\sum_{k=1}^{N}
\exp(z_k)}.
\end{equation}
When the scores $z_j$ are large, the exponentials $\exp(z_j)$ can overflow, leading to numerical errors.
A common solution is to subtract the row-wise maximum $m = \max_j z_j$ before exponentiation:
\begin{equation}
\text{softmax}(z_j) = \frac{\exp(z_j - m)}{\sum_{k=1}^{N}
\exp(z_k - m)}.
\end{equation}
This is valid because softmax is shift-invariant: for any constant $c$,
\begin{equation}
\frac{\exp(z_j - c)}{\sum_{k} \exp(z_k - c)} =
\frac{\exp(z_j) \cdot \exp(-c)}{\sum_{k} \exp(z_k) \cdot
\exp(-c)} = \frac{\exp(z_j)}{\sum_{k} \exp(z_k)}.
\end{equation}
Subtracting the maximum ensures that the largest exponent is $\exp(0) = 1$, preventing overflow while producing identical results.

\paragraph{Log-Sum-Exp (LSE).}
The denominator of softmax can be compactly represented using the Log-Sum-Exp (LSE).
For a score vector $\mathbf{z} = [z_1, z_2, \ldots, z_N]$, LSE is defined as:
\begin{equation}
\text{LSE}(\mathbf{z}) = m + \log \sum_{j=1}^{N}
\exp(z_j - m), \quad \text{where } m = \max_j z_j.
\label{eq:lse}
\end{equation}
With LSE, each softmax weight $w_j$ for the $j$-th element can be recovered as:
\begin{equation}
w_j = \exp(z_j - \text{LSE}(\mathbf{z})).
\end{equation}
LSE serves as the global normalization factor in log-space.
It plays a central role in FlashAttention's incremental computation, as we describe next.

\paragraph{Online Softmax and Incremental Computation.}
Standard softmax requires two passes over the data: one to compute the maximum and denominator, and another to normalize.
This is impractical when the full score vector $\mathbf{z}$ is not available in memory, as is the case in FlashAttention where scores are computed block by block.

Online softmax solves this by \textbf{simultaneously accumulating the numerator and denominator of softmax} as each block is processed, \textbf{deferring the final division to the end}.
Suppose we have already processed blocks $1, \ldots, t-1$ with running statistics $m^{(t-1)}$ and $\ell^{(t-1)}$.
When a new block $t$ with local scores $\mathbf{z}^{(t)}$ arrives, the update proceeds as follows:
\begin{align}
m^{(t)}_{\text{local}} &= \max_j z^{(t)}_j,
\\
m^{(t)} &= \max(m^{(t-1)},\; m^{(t)}_{\text{local}}),
\\
\ell^{(t)} &= \ell^{(t-1)} \cdot
\exp(m^{(t-1)} - m^{(t)}) + \sum_j
\exp(z^{(t)}_j - m^{(t)}).
\end{align}
The key insight is that when a new maximum is encountered, all previously accumulated sums must be rescaled by $\exp(m^{(t-1)} - m^{(t)})$ to remain on the same numerical basis.
This rescaling is again justified by the shift-invariance of softmax.

Similarly, the partial attention output $\mathbf{o}$ is rescaled at each step:
\begin{equation}
\mathbf{o}^{(t)} = \mathbf{o}^{(t-1)} \cdot
\exp(m^{(t-1)} - m^{(t)}) + \sum_j
\exp(z^{(t)}_j - m^{(t)}) \cdot \mathbf{v}^{(t)}_j.
\end{equation}
After all blocks have been processed, the final output is normalized by the accumulated sum:
\begin{equation}
\mathbf{o}_{\text{final}} = \frac{\mathbf{o}^{(T)}}
{\ell^{(T)}},
\end{equation}
and the final LSE is computed as:
\begin{equation}
\text{LSE} = m^{(T)} + \log \ell^{(T)}.
\end{equation}

\paragraph{Why LSE Is Only Available After Completion.}
\label{par:lse_availability}
The running maximum $m^{(t)}$ may change as new blocks are processed.
Each such change triggers a rescaling of all previously accumulated results.
Therefore, the true global maximum and the true LSE are only known after the last block has been processed.
Any block weight computed during the loop using intermediate statistics will be based on an incomplete normalization factor, and thus will not be precisely normalized.

This is exactly the challenge that GSRS's two-phase correction addresses (Section~\ref{sec:precise}): Phase 1 computes block weights using a cached LSE as an approximate reference during the kernel, and Phase 2 corrects them with the true LSE once the kernel completes.

\paragraph{Summary.} Algorithm~\ref{alg:flash_attn} summarizes the complete FlashAttention procedure described above.

\begin{algorithm}[H]
\small
\setlength{\abovecaptionskip}{0pt} 
\setlength{\belowcaptionskip}{0pt} 
\caption{FlashAttention with Online Softmax}
\label{alg:flash_attn}
\begin{algorithmic}[1]
\Require $\mathbf{Q}, \mathbf{K}, \mathbf{V}$: query, key,
value matrices; $B$: block size; $N_b = L/B$: number of
blocks
\Ensure $\mathbf{O}$: attention output; $\text{LSE}$:
log-sum-exp values
\For{each query block $u = 1, \ldots, N_b$}
    \State Initialize $\mathbf{o}_i \gets 0$,
    $\ell_i \gets 0$, $m_i \gets -\infty$ for each row
    $i \in \text{Block } u$
    \For{each key-value block $v = 1, \ldots, N_b$}
        \State $\mathbf{z}^{(v)} \gets \mathbf{Q}_u
        \mathbf{K}_v^\top / \sqrt{d}$
        \Comment{Local scores}
        \State $m_{\text{local}} \gets \max_j z^{(v)}_j$
        \State $m_i^{\text{new}} \gets \max(m_i,\;
        m_{\text{local}})$
        \Comment{Update maximum}
        \State $\ell_i \gets \ell_i \cdot \exp(m_i -
        m_i^{\text{new}}) + \sum_j \exp(z^{(v)}_j -
        m_i^{\text{new}})$
        \Comment{Denominator}
        \State $\mathbf{o}_i \gets \mathbf{o}_i \cdot
        \exp(m_i - m_i^{\text{new}}) + \sum_j
        \exp(z^{(v)}_j - m_i^{\text{new}}) \cdot
        \mathbf{v}^{(v)}_j$
        \Comment{Numerator}
        \State $m_i \gets m_i^{\text{new}}$
    \EndFor
    \State $\mathbf{o}_i \gets \mathbf{o}_i / \ell_i$
    \Comment{Final normalization}
    \State $\text{LSE}_i \gets m_i + \log \ell_i$
    \Comment{Compute LSE}
\EndFor
\State \Return $\mathbf{O}$, $\text{LSE}$
\end{algorithmic}
\end{algorithm}

\newpage
\subsection{Equivalence to the Block Importance Definition}
\label{sec:app_block_importance_proof}

We show that the two-phase evaluation recovers the block importance defined in Eq.~\eqref{eq:block_importance}.
For query block $u$, let $\mathcal{C}$ denote the evaluated key blocks and $\mathcal{K}_{\mathcal{C}}$ their key tokens.
This set contains all key blocks at the first search and the candidates defined by $\mathcal{S}_1$ at subsequent searches.
For each query row $i\in u$ and key block $v\in\mathcal{C}$, Phase 1 stores
\begin{equation}
    G_{iv}=m_{iv}+\log\sum_{j\in v}\exp(z_{ij}-m_{iv}),
    \qquad m_{iv}=\max_{j\in v}z_{ij}.
\end{equation}
Since $m_{iv}$ is constant across keys within block $v$, it cancels when the expression is expanded:
\begin{equation}
\begin{aligned}
    G_{iv}
    &=m_{iv}+\log\left(\exp(-m_{iv})\sum_{j\in v}\exp(z_{ij})\right) \\
    &=m_{iv}-m_{iv}+\log\sum_{j\in v}\exp(z_{ij}) \\
    &=\log\sum_{j\in v}\exp(z_{ij}).
\end{aligned}
\end{equation}
Subtracting the block maximum before exponentiation provides numerical stability while preserving the block log-sum-exp.

After all candidate blocks have been processed, FlashAttention provides the current row-wise LSE $L_i=\log\sum_{j\in\mathcal{K}_{\mathcal{C}}}\exp(z_{ij})$.
Substituting the expression for $G_{iv}$ into Phase 2 gives
\begin{equation}
\begin{aligned}
    W_{uv}
    &=\frac{1}{B}\sum_{i\in u}\exp(G_{iv}-L_i) \\
    &=\frac{1}{B}\sum_{i\in u}\exp(-L_i)\sum_{j\in v}\exp(z_{ij}) \\
    &=\frac{1}{B}\sum_{i\in u}\sum_{j\in v}\exp(z_{ij}-L_i) \\
    &=\frac{1}{B}\sum_{i\in u}\sum_{j\in v}w_{ij}.
\end{aligned}
\end{equation}
This is the block importance in Eq.~\eqref{eq:block_importance}, with attention normalized over the evaluated key tokens.
Consequently, $\sum_{v\in\mathcal{C}}W_{uv}=1$, providing normalized block weights for Top-p selection without cached LSE values.

\newpage
\subsection{Fused Precise Block Importance Evaluation}
\label{sec:app_fused_eval}

Algorithm~\ref{alg:fused_eval} presents the complete procedure of the fused block importance evaluation described in Section~\ref{sec:precise}.
Phase 1 is executed within the FlashAttention kernel, while Phase 2 is a lightweight post-kernel operation.

\begin{algorithm}[H]
    \small
    \setlength{\abovecaptionskip}{0pt}
    \setlength{\belowcaptionskip}{0pt}
    \caption{Fused Precise Block Importance Evaluation without Cached LSE}
    \label{alg:fused_eval}
    \begin{algorithmic}[1]
        \Require $\mathbf{Q},\mathbf{K},\mathbf{V}$: query, key, value matrices; $\mathcal{S}$: search candidate mask; $B$: block size
        \Ensure $\mathbf{O}$: attention output; $L$: current row-wise LSE; $\mathbf{W}$: block importance matrix
        \State Initialize $G_{iv}\gets-\infty$ and $W_{uv}\gets0$
        \Statex \textit{// Phase 1: In-kernel attention and block log-sum-exp}
        \For{each query block $u$ in parallel}
            \State Initialize $\mathbf{o}_i\gets\mathbf{0}$, $\ell_i\gets0$, $m_i\gets-\infty$ for $i\in u$
            \For{each key-value block $v\in\mathcal{S}[u]$}
                \State $\mathbf{Z}_{uv}\gets\mathbf{Q}_u\mathbf{K}_v^\top/\sqrt{d}$
                \State $m_{iv}\gets\max_{j\in v}z_{ij}$, $P_{ij}\gets\exp(z_{ij}-m_{iv})$
                \State $r_{iv}\gets\sum_{j\in v}P_{ij}$
                \State $m_i'\gets\max(m_i,m_{iv})$
                \State $\alpha_i\gets\exp(m_i-m_i')$, $\beta_i\gets\exp(m_{iv}-m_i')$
                \State $\ell_i\gets\alpha_i\ell_i+\beta_i r_{iv}$
                \State $\mathbf{o}_i\gets\alpha_i\mathbf{o}_i+\beta_i\sum_{j\in v}P_{ij}\mathbf{v}_j$
                \State $m_i\gets m_i'$
                \State Store $G_{iv}\gets m_{iv}+\log r_{iv}$
            \EndFor
            \State $L_i\gets m_i+\log\ell_i$, $\mathbf{O}_i\gets\mathbf{o}_i/\ell_i$
            \State Write $\mathbf{O}_u$ and $L_i$ for $i\in u$
        \EndFor
        \Statex \textit{// Phase 2: Post-kernel block importance recovery}
        \For{each query block $u$ and candidate block $v\in\mathcal{S}[u]$}
            \State $W_{uv}\gets\frac{1}{B}\sum_{i\in u}\exp(G_{iv}-L_i)$
        \EndFor
        \State \Return $\mathbf{O}$, $L$, $\mathbf{W}$
    \end{algorithmic}
\end{algorithm}
For fully masked block rows, we set $P_{ij}=0$, $r_{iv}=0$, and $G_{iv}=-\infty$; the corresponding scale factor $\beta_i$ is zero.
The history scale $\alpha_i$ is zero when no valid key has yet been processed.
Rows with no valid keys return $\mathbf{O}_i=\mathbf{0}$ and $L_i=-\infty$ and contribute zero to block importance.

\paragraph{Memory Overhead.}
\label{par:peak_memory}
The primary memory overhead of GSRS comes from storing the row-level weight sums $\hat{S}_{iv}$ during the fused evaluation kernel.
For a sequence of length $L$ with block size $B$ and $H$ attention heads, the storage requirement is $H \times \frac{L}{B} \times \frac{L}{B} \times B = H \times \frac{L^2}{B}$ elements.
Compared to materializing the full attention matrix of size $H \times L \times L$, this reduces memory by a factor of $B$.
For example, with $L = 115200$, $B = 128$, and $H = 24$ on HunyuanVideo, the storage is approximately 4.6 GB in bfloat16.
As shown in Table~\ref{tab:memory}, GSRS achieves slightly lower peak memory than full attention despite the additional storage for $\hat{S}_{iv}$, indicating that the memory overhead of block importance evaluation does not increase the overall memory footprint in practice.

\begin{table}[H]
    \centering
    \vspace{20pt}
    \caption{Peak GPU memory usage during inference on HunyuanVideo (single A800, 125 frames, 720p).}
    \label{tab:memory}
    \begin{tabular}{lc}
    \toprule
    \textbf{Configuration} & \textbf{Peak Memory (GB)} \\
    \midrule
    Full Attention & 65.60 \\
    GSRS & 62.65 \\
\bottomrule

\vspace{10pt}
\end{tabular}
\end{table}

\begin{figure}[H]
    \centering
    \includegraphics[width=\textwidth]{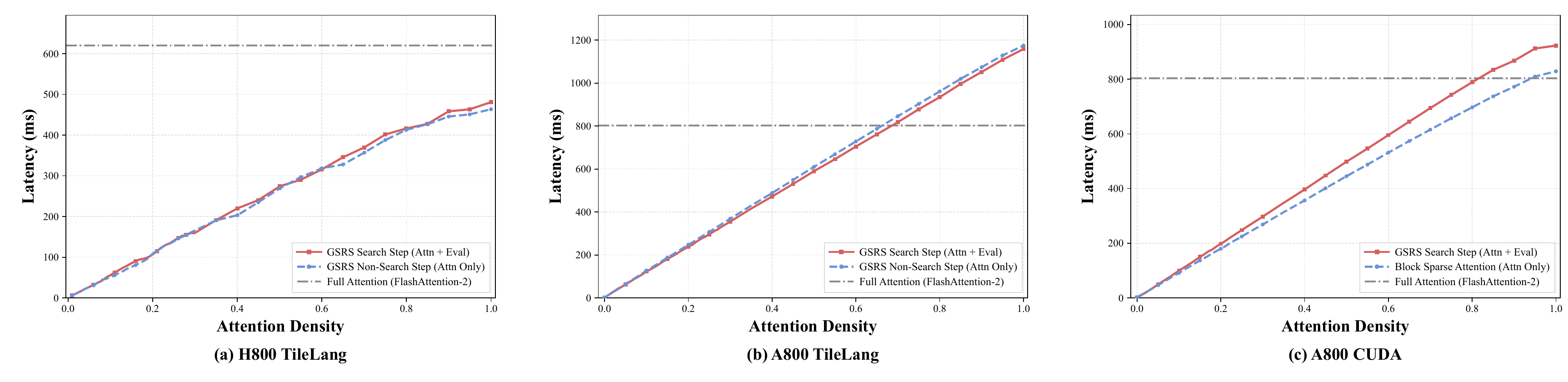}
    \caption{Per-step attention kernel latency at varying density levels ($L=115200, H=24, d=128$).
    }
    \label{fig:kernel_latency}
\end{figure}

\paragraph{Kernel Latency Analysis.}
Beyond memory efficiency, we evaluate the overhead of fused block importance evaluation using TileLang \cite{tilelang2024} implementations on H800 and A800 GPUs, as well as a CUDA implementation on A800.
Figure~\ref{fig:kernel_latency} compares attention kernel latency with and without fused block statistics computation at varying density levels.
Search steps perform both attention computation and block importance evaluation, while non-search steps perform attention computation only.
For the CUDA implementation, we use the attention-only kernels from Block Sparse Attention \cite{guo2024block} as the baseline.
The small latency gap across the tested implementations indicates that collecting block log-sum-exp statistics $G_{iv}$ adds limited overhead to attention computation.

\subsection{Top-p Block Selection}
\label{sec:app_top_p}

Algorithm~\ref{alg:top_p} describes the \textit{Top-p} block selection procedure used to construct the sparse mask from the block importance matrix $\mathbf{W}$.

\begin{algorithm}[H]
    \caption{Top-p Block Selection}
    \label{alg:top_p}
    \begin{algorithmic}[1]
    \Require $\mathbf{W}$: block importance matrix of shape
    $[H, N_q, N_k]$; $p$: \textit{Top-p} threshold
    \Ensure $\mathcal{S}$: selected block mask
    \For{each head $h$, each query block $u$}
        \State Sort $\{W_{uv}\}_{v=1}^{N_k}$ in descending order,
        yielding permutation $\pi$
        \State Compute cumulative sum $C_v = \sum_{k=1}^{v}
        W_{u,\pi(k)}$
        \State Find smallest $v^*$ such that $C_{v^*} \geq p$
        \State $\mathcal{S}[h, u] \gets \{\pi(1), \ldots, \pi(v^*)\}$
    \EndFor
    \State \Return $\mathcal{S}$
    \end{algorithmic}
\end{algorithm}

\newpage
\subsection{GSRS Pipeline}
\label{sec:app_pipeline}

Algorithm~\ref{alg:GSRS_pipeline} presents the complete GSRS inference pipeline, integrating warmup, fused evaluation, Growing Sparsity and Reduced Search, and mask reuse across denoising steps.

\begin{algorithm}[H]
\caption{GSRS Pipeline}
\label{alg:GSRS_pipeline}
\begin{algorithmic}[1]
\Require Per-step $\mathbf{Q}_t,\mathbf{K}_t,\mathbf{V}_t$; $N_w$: warmup steps; $\mathcal{T}=\{t_1,\ldots,t_n\}$: search steps with $t_1\geq N_w$; $\{p_1,\ldots,p_n\}$: Top-p thresholds; $B$: block size
\Ensure $\mathbf{O}_t$: attention output at each step
\State $\mathcal{S}_{\mathrm{latest}}\gets\mathcal{S}_{\mathrm{full}}$
\For{each denoising step $t=1,\ldots,T$}
    \If{$t=t_1$}
        \State $\mathbf{O}_t,\_,\mathbf{W}\gets\text{FusedEval}(\mathbf{Q}_t,\mathbf{K}_t,\mathbf{V}_t,\mathcal{S}_{\mathrm{full}},B)$
        \Comment{Full attention and first search}
        \State $\mathcal{S}_1\gets\text{Top-p}(\mathbf{W},p_1)$
        \State $\mathcal{S}_{\mathrm{latest}}\gets\mathcal{S}_1$
    \ElsIf{$t=t_k\in\mathcal{T}$ for $k\geq2$}
        \State $\mathbf{O}_t,\_,\mathbf{W}\gets\text{FusedEval}(\mathbf{Q}_t,\mathbf{K}_t,\mathbf{V}_t,\mathcal{S}_1,B)$
        \Comment{Re-evaluate all candidates in $\mathcal{S}_1$}
        \State $\mathcal{S}_k\gets\text{Top-p}(\mathbf{W},p_k)$
        \State $\mathcal{S}_{\mathrm{latest}}\gets\mathcal{S}_k$
    \ElsIf{$t\leq N_w$ or $t<t_1$}
        \State $\mathbf{O}_t\gets\text{FullAttention}(\mathbf{Q}_t,\mathbf{K}_t,\mathbf{V}_t)$
        \Comment{Warmup and steps before the first search}
    \Else
        \State $\mathbf{O}_t\gets\text{SparseAttention}(\mathbf{Q}_t,\mathbf{K}_t,\mathbf{V}_t,\mathcal{S}_{\mathrm{latest}})$
        \Comment{Use the latest selected mask}
    \EndIf
\EndFor
\end{algorithmic}
\end{algorithm}
The first search computes full attention and can coincide with the last warmup step.
The current LSE is used within FusedEval to recover block importance and is not cached across denoising steps.

\subsection{Token Reordering via Hilbert Curve}
\label{sec:app_hilbert}
\begin{figure}[H]
\centering
\includegraphics[width=0.65\textwidth]{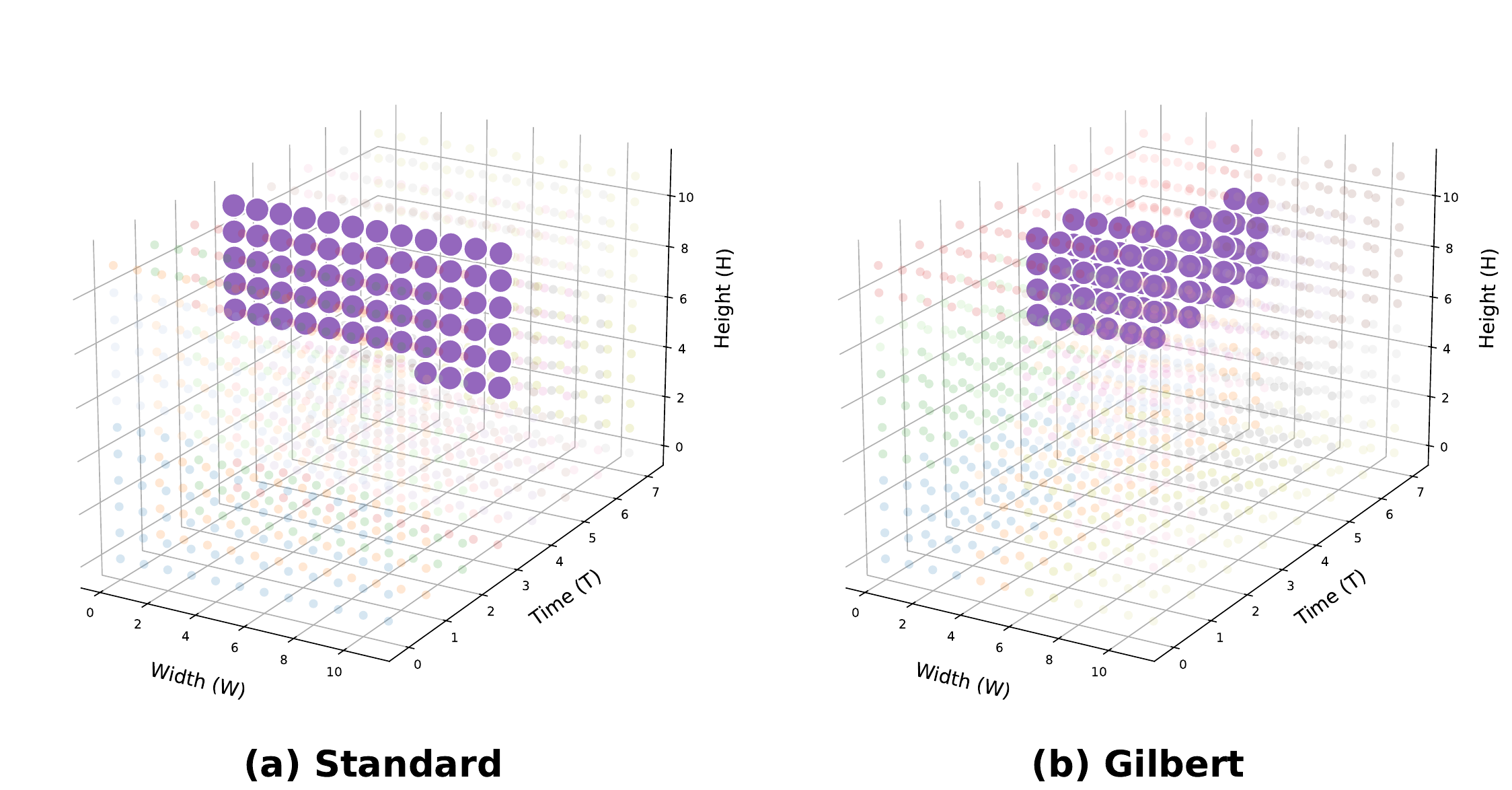}
\caption{Token distribution within a single block under different ordering strategies.
(a) Standard row-major ordering groups tokens from a flat spatial-temporal slab.
(b) Gilbert curve ordering groups tokens from a compact spatial-temporal region, better preserving spatial-temporal adjacency.}
\label{fig:hilbert}
\end{figure}

Flattening a 3D video tensor into a 1D sequence in row-major order can separate spatiotemporally adjacent tokens, weakening locality within fixed-size blocks (Figure~\ref{fig:hilbert}(a)).
We therefore reorder tokens using the Generalized Hilbert curve \cite{cerveny2025gilbert}, which preserves spatial-temporal locality and groups nearby tokens into compact blocks (Figure~\ref{fig:hilbert}(b)).
This helps block sparse attention capture local interactions.

Following prior work \cite{cerveny2025gilbert,zhang2025fast,zhangtraining}, we apply this reordering in all experiments.

\subsection{GSRS Runtime Breakdown}
\label{sec:runtime_breakdown}
\begin{figure}[h]
\centering
\includegraphics[width=0.9\textwidth]{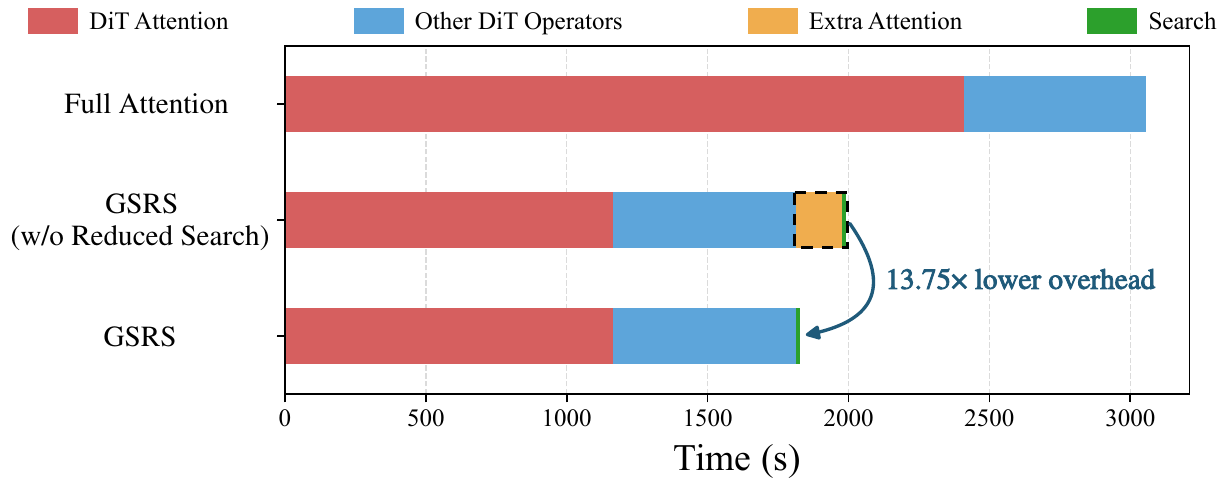}
\caption{\textbf{Runtime breakdown on HunyuanVideo.}
Removing reduced search introduces substantial extra attention and search overhead.
GSRS with reduced search achieves 13.75$\times$ lower overhead compared to the variant without it.}
\label{fig:runtime_breakdown}
\end{figure}

Figure~\ref{fig:runtime_breakdown} presents the runtime breakdown of GSRS on HunyuanVideo.
Without reduced search, each search step evaluates all blocks in the attention matrix rather than restricting to the baseline mask.
Evaluating all blocks not only increases the search time at each search step, but also retains more blocks due to the larger search space, leading to additional attention computation time (shown as Extra Attention in the figure).
The search overhead itself (shown as Search) is dominated by intermediate matrix I/O, rescaling, and averaging, as the efficient kernel fusion described in Section~\ref{sec:precise} minimizes the computational cost.
We define the total overhead of removing reduced search as the sum of Extra Attention and Search.
As shown in the figure, reduced search lowers this total overhead by 13.75$\times$, confirming that restricting searches to the baseline mask is essential for practical efficiency.

\section{Details about VBench Result}
    \label{sec:vbench_result}
    \begin{table*}[h]
    \centering
    \renewcommand{\arraystretch}{1.3}
    \caption{\textbf{Detailed VBench Quality dimensions.}
    Full breakdown of VBench Quality Score across all 7 dimensions for Wan 2.1 and HunyuanVideo.
    \textbf{Bold} and \underline{underline} denote the best and second best results.}
    \label{tab:vbench_full}
    \resizebox{\textwidth}{!}{
    \begin{tabular}{ll!{\color{gray!40}\vrule width 0.4pt}ccccccc!{\color{gray!40}\vrule width 0.4pt}c}
    \toprule
    \textbf{Model} & \textbf{Method}
    & \makecell{\textbf{Subject} \\ \textbf{Consistency}}
    & \makecell{\textbf{Background} \\ \textbf{Consistency}}
    & \makecell{\textbf{Temporal} \\ \textbf{Flickering}}
    & \makecell{\textbf{Motion} \\ \textbf{Smoothness}}
    & \makecell{\textbf{Aesthetic} \\ \textbf{Quality}}
    & \makecell{\textbf{Imaging} \\ \textbf{Quality}}
    & \makecell{\textbf{Dynamic} \\ \textbf{Degree}}
    & \makecell{\textbf{VBench-Q}} \\
    \midrule
    \multirow{7}{*}{\textbf{Wan 2.1}}
    & \cellcolor{gray!9}Full Attention    & \cellcolor{gray!9}98.26\% & \cellcolor{gray!9}98.25\% & \cellcolor{gray!9}99.08\% & \cellcolor{gray!9}99.21\% & \cellcolor{gray!9}63.07\% & \cellcolor{gray!9}73.29\% & \cellcolor{gray!9}33.52\% & \cellcolor{gray!9}83.75\% \\
    & SpargeAttn \cite{zhang2025spargeattention} & 94.88\% & 95.49\% & 99.07\% & 99.15\% & 61.73\% & \underline{72.80\%} & 32.95\% & 82.21\% \\
    & SVG \cite{xi2025sparse}              & 97.74\% & \underline{97.90\%} & \underline{99.27\%} & \textbf{99.43\%} & \textbf{63.18\%} & 71.99\% & \textbf{34.65\%} & \underline{83.69\%} \\
    & SVG 2 \cite{yangsparse}              & 97.80\% & 97.47\% & 99.26\% & 99.38\% & 61.87\% & 72.23\% & 31.81\% & 83.19\% \\
    & SVOO \cite{svoo}                     & \underline{97.89\%} & 97.76\% & \textbf{99.30\%} & \underline{99.40\%} & 61.63\% & 72.42\% & 32.38\% & 83.33\% \\

    & \cellcolor{blue!8}\textbf{GSRS} & \cellcolor{blue!8}\textbf{98.21\%} & \cellcolor{blue!8}\textbf{98.22\%} & \cellcolor{blue!8}99.07\% & \cellcolor{blue!8}99.19\% & \cellcolor{blue!8}\underline{63.09\%} & \cellcolor{blue!8}\textbf{73.22\%} & \cellcolor{blue!8}\underline{33.52\%} & \cellcolor{blue!8}\textbf{83.72\%} \\
    \midrule
    \multirow{7}{*}{\textbf{HunyuanVideo}}
    & \cellcolor{gray!9}Full Attention    & \cellcolor{gray!9}96.82\% & \cellcolor{gray!9}97.07\% & \cellcolor{gray!9}98.62\% & \cellcolor{gray!9}99.51\% & \cellcolor{gray!9}56.40\% & \cellcolor{gray!9}70.71\% & \cellcolor{gray!9}61.93\% & \cellcolor{gray!9}83.98\% \\
    & SpargeAttn \cite{zhang2025spargeattention} & 96.55\% & 96.68\% & 98.52\% & 99.48\% & \underline{56.25\%} & \underline{70.65\%} & \textbf{63.06\%} & \underline{83.85\%} \\
    & SVG \cite{xi2025sparse}              & \underline{96.71\%} & \underline{96.85\%} & 98.70\% & \textbf{99.52\%} & 55.60\% & 68.98\% & \underline{57.38\%} & 83.21\% \\
    & SVG 2 \cite{yangsparse}              & 96.60\% & 96.05\% & \textbf{98.77\%} & 99.49\% & 54.05\% & 66.50\% & 55.11\% & 82.24\% \\
    & SVOO \cite{svoo}                     & \textbf{96.88\%} & 96.39\% & \underline{98.73\%} & \underline{99.51\%} & 55.05\% & 69.19\% & 55.68\% & 82.97\% \\


    & \cellcolor{blue!8}\textbf{GSRS}  & \cellcolor{blue!8}96.66\% & \cellcolor{blue!8}\textbf{96.91\%} & \cellcolor{blue!8}98.58\% & \cellcolor{blue!8}99.49\% & \cellcolor{blue!8}\textbf{56.51\%} & \cellcolor{blue!8}\textbf{70.79\%} & \cellcolor{blue!8}\textbf{63.06\%} & \cellcolor{blue!8}\textbf{84.01\%} \\
    \bottomrule
    \end{tabular}
    }
\end{table*}

\section{Effect of Warmup Length}
\label{sec:warmup_length}
Table~\ref{tab:warmup_ablation} shows that increasing the number of warmup steps improves similarity to full attention at the cost of higher latency.
We use $N_w=10$ to balance similarity and speedup.
Meanwhile, VBench-Q remains between 83.71\% and 84.01\% across warmup settings (Table~\ref{tab:warmup_vbench_full}), indicating that generation quality is largely preserved even with $N_w=1$.
\begin{table}[h]
    \centering
    \small
    \caption{Ablation on the number of warmup steps $N_w$ on HunyuanVideo.}
    \label{tab:warmup_ablation}
    \begin{tabular}{lcccc}
    \toprule
    $N_w$ & PSNR$\uparrow$ & SSIM$\uparrow$ & LPIPS$\downarrow$ & Latency(s)$\downarrow$ \\
    \midrule
    1     & 24.60         & 0.835         & 0.258                & 1680                  \\
    3     & 27.63         & 0.899         & 0.203                & 1728                  \\
    5     & 29.39         & 0.925         & 0.180                & 1746                  \\
    8     & 31.00         & 0.944         & 0.161                & 1785                  \\
    10    & 32.37         & 0.957         & 0.138                & 1825                  \\
    15    & 33.79         & 0.966         & 0.134                & 1992                  \\
    \bottomrule
    \end{tabular}
\end{table}

\begin{table*}[h]
    \centering
    \small
    \renewcommand{\arraystretch}{1.3}
    \caption{\textbf{Detailed VBench Quality dimensions under different warmup steps.}
    Full breakdown of VBench Quality Score across all 7 dimensions on HunyuanVideo.}
    \label{tab:warmup_vbench_full}
    \resizebox{\textwidth}{!}{
    \begin{tabular}{c!{\color{gray!40}\vrule width 0.4pt}ccccccc!{\color{gray!40}\vrule width 0.4pt}c}
    \toprule
    \makecell{\textbf{Warmup} \\ \textbf{Steps}}
    & \makecell{\textbf{Subject} \\ \textbf{Consistency}}
    & \makecell{\textbf{Background} \\ \textbf{Consistency}}
    & \makecell{\textbf{Temporal} \\ \textbf{Flickering}}
    & \makecell{\textbf{Motion} \\ \textbf{Smoothness}}
    & \makecell{\textbf{Aesthetic} \\ \textbf{Quality}}
    & \makecell{\textbf{Imaging} \\ \textbf{Quality}}
    & \makecell{\textbf{Dynamic} \\ \textbf{Degree}}
    & \makecell{\textbf{VBench-Q}} \\
    \midrule
    \textbf{1}  & 96.64\% & 96.44\% & 98.55\% & 99.49\% & 56.02\% & 70.66\% & 63.06\% & 83.80\% \\
    \textbf{3}  & 96.66\% & 96.50\% & 98.59\% & 99.49\% & 55.89\% & 70.86\% & 61.36\% & 83.71\% \\
    \textbf{5}  & 96.68\% & 96.50\% & 98.60\% & 99.49\% & 56.03\% & 70.81\% & 61.93\% & 83.78\% \\
    \textbf{8}  & 96.66\% & 96.52\% & 98.61\% & 99.50\% & 55.94\% & 70.82\% & 62.50\% & 83.81\% \\
    \textbf{10} & 96.66\% & 96.91\% & 98.58\% & 99.49\% & 56.51\% & 70.79\% & 63.06\% & 84.01\% \\
    \textbf{15} & 96.67\% & 96.49\% & 98.62\% & 99.50\% & 55.89\% & 70.78\% & 62.50\% & 83.80\% \\
    \bottomrule
    \end{tabular}
    }
\end{table*}

\section{PSNR Distribution}
\label{sec:psnr_distribution}
\begin{figure}[H]
    \centering
    \includegraphics[width=1.0\textwidth]{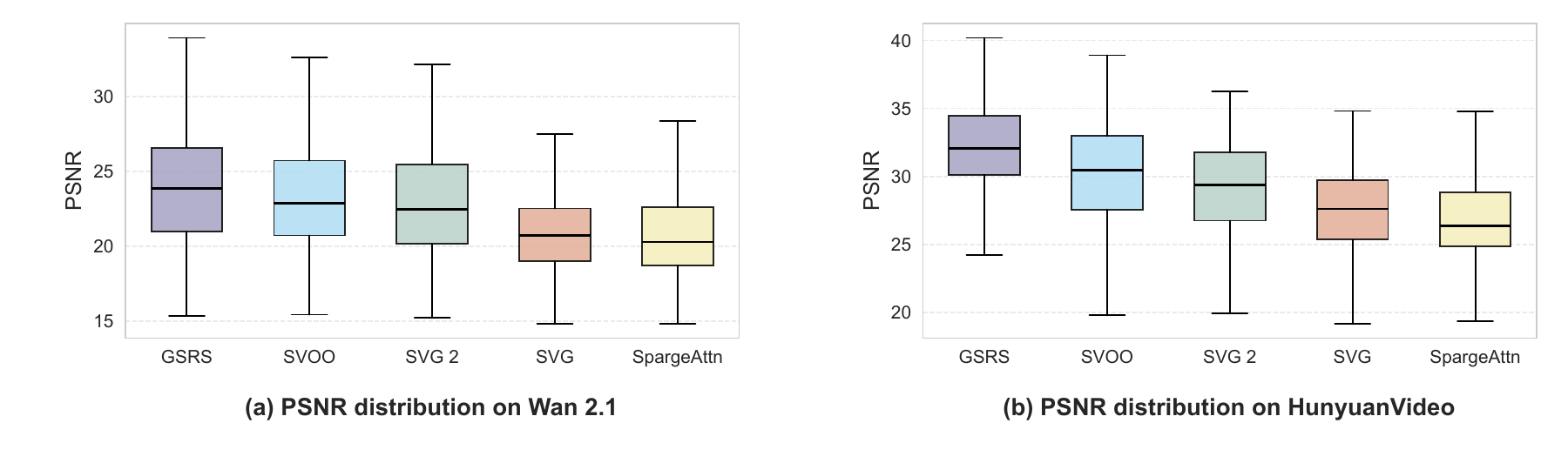}
    \vspace{-10pt}
    \caption{\textbf{PSNR distribution across VBench prompts.}
    GSRS achieves the highest median PSNR on both Wan2.1 and HunyuanVideo.}
    \label{fig:psnr_distribution}
\end{figure}

Figure~\ref{fig:psnr_distribution} compares PSNR distributions across VBench prompts on Wan2.1 and HunyuanVideo.
GSRS achieves the highest median PSNR on both models, consistent with the average scores in Table~\ref{tab:main_results}.
Its lower and upper quartiles are also higher than those of all baselines, showing improved similarity to full attention across diverse prompts.

\section{Performance Across Random Seeds}
\label{sec:random_seeds}
We evaluate GSRS on HunyuanVideo using five random seeds and report the mean and standard deviation of PSNR, SSIM, LPIPS, and VBench-Q.
PSNR, SSIM, and LPIPS are measured against full-attention outputs generated with the same seed.

\begin{table}[htbp]
    \centering
    \caption{GSRS performance across five random seeds on HunyuanVideo.
    Values are reported as mean $\pm$ standard deviation.}
    \label{tab:random_seeds}
    \begin{tabular}{lcccc}
        \toprule
        Model        & PSNR$\uparrow$      & SSIM$\uparrow$         & LPIPS$\downarrow$      & VBench-Q$\uparrow$ \\
        \midrule
        HunyuanVideo & $31.35 \pm 0.97$    & $0.9519 \pm 0.0067$    & $0.147 \pm 0.011$      & $84.43 \pm 0.65$ \\
        \bottomrule
    \end{tabular}
\end{table}

\section{Quality-Efficiency Trade-off}
\label{sec:top_p_configurations}
We evaluate the quality-efficiency trade-off of GSRS on HunyuanVideo by varying the Top-p thresholds across search steps.
Table~\ref{tab:top_p_configurations} reports PSNR, SSIM, LPIPS, and VBench-Q alongside inference latency and speedup.

\begin{table}[htbp]
    \centering
    \caption{GSRS performance under different Top-p configurations on HunyuanVideo.}
    \label{tab:top_p_configurations}
    \begin{tabular}{lcccccc}
        \toprule
        Top-p configuration        & PSNR$\uparrow$ & SSIM$\uparrow$ & LPIPS$\downarrow$ & VBench-Q$\uparrow$   & Latency$\downarrow$  & Speedup$\uparrow$ \\
        \midrule
        \{0.95, 0.95, 0.95, 0.95\} & 32.55          & 0.959          & 0.136             & 84.89                & 1969s                & 1.55$\times$               \\
        \{0.93, 0.93, 0.93, 0.93\} & 31.20          & 0.947          & 0.153             & 84.90                & 1852s                & 1.65$\times$               \\
        \{0.90, 0.90, 0.90, 0.90\} & 30.28          & 0.937          & 0.166             & 84.89                & 1774s                & 1.72$\times$               \\
        \{0.88, 0.88, 0.88, 0.88\} & 29.63          & 0.930          & 0.170             & 84.95                & 1749s                & 1.75$\times$               \\
        \midrule
        \{0.95, 0.80, 0.85, 0.90\} & 31.64          & 0.951          & 0.143             & 84.98                & 1850s                & 1.65$\times$               \\
        \{0.90, 0.90, 0.95, 0.95\} & 30.40          & 0.939          & 0.163             & 84.87                & 1792s                & 1.70$\times$               \\
        \{0.95, 0.95, 0.90, 0.90\} & 32.38          & 0.957          & 0.141             & 84.85                & 1907s                & 1.60$\times$               \\
        \{0.90, 0.85, 0.80, 0.75\} & 30.00          & 0.933          & 0.174             & 84.85                & 1719s                & 1.78$\times$               \\
        \{0.85, 0.80, 0.75, 0.70\} & 28.95          & 0.920          & 0.190             & 84.78                & 1631s                & 1.87$\times$               \\
        \rowcolor{blue!8}
        \{0.95, 0.90, 0.85, 0.80\} & 31.96          & 0.953          & 0.141             & 84.99                & 1825s                & 1.68$\times$               \\
        \bottomrule
    \end{tabular}
\end{table}

\section{Inference Speed on NVIDIA H800}
\label{sec:h800_speed}
We compare the inference latency and speedup of GSRS and compared sparse attention methods on a single NVIDIA H800 GPU.
Speedup is measured relative to full attention on the same GPU.

\begin{table}[htbp]
    \centering
    \caption{Inference latency and speedup on a single NVIDIA H800 GPU.}
    \label{tab:h800_speed}
    \begin{tabular}{lcccc}
        \toprule
        & \multicolumn{2}{c}{Wan2.1}
        & \multicolumn{2}{c}{HunyuanVideo} \\
        \cmidrule(lr){2-3}\cmidrule(lr){4-5}
        Method
        & Latency $\downarrow$ & Speedup$\uparrow$
        & Latency $\downarrow$ & Speedup$\uparrow$ \\
        \midrule
        \rowcolor{gray!9}
        Full Attention & 2573s & 1.00$\times$ & 2265s & 1.00$\times$ \\
        SpargeAttn     & 1934s & 1.33$\times$           & 1497s & 1.51$\times$           \\
        SVG            & 1796s & 1.43$\times$           & 1319s & 1.72$\times$           \\
        SVG2           & \underline{1735s} & \underline{1.48$\times$}           & 1168s & 1.94$\times$           \\
        SVOO           & 1759s & 1.46$\times$           & \underline{1143s} & \underline{1.98$\times$}           \\
        \rowcolor{blue!8}
        \textbf{GSRS}  & \textbf{1443s} & \textbf{1.78$\times$}           & \textbf{1026s} & \textbf{2.21$\times$}           \\
        \bottomrule
    \end{tabular}
\end{table}

\newpage
\section{Limitations}
\label{sec:limitations}
GSRS has two limitations that warrant future investigation.
First, at least one full-attention warmup step is required to obtain the initial LSE for subsequent block importance evaluation.
Although the warmup overhead is small relative to the total inference time, eliminating this requirement would further improve efficiency.
Second, GSRS is designed and evaluated specifically for DiT-based video generation models.
Whether the observed attention dynamics and the proposed mechanisms generalize to other architectures such as large language models remains an open question for future work.

\section{Broader Impacts}
\label{sec:broader_impacts}
This work improves the efficiency of video diffusion models, reducing computational cost and energy consumption for high-quality video generation.
By lowering the hardware requirements, GSRS makes video generation more accessible to researchers and creators with limited computational resources, potentially democratizing access to this technology.

However, improved efficiency may also lower the barrier to generating realistic synthetic videos at scale, which could be misused for producing misleading or deceptive content such as deepfakes.
The ability to generate videos faster and cheaper amplifies both beneficial and harmful applications.
We encourage responsible deployment of this technology and recommend combining video generation systems with safeguards such as content watermarking and synthetic media detection to mitigate potential misuse.

\section{Detailed Comparison with AdaSpa and LiteAttention}
\label{sec:comparison}

\paragraph{Search and block selection.}
AdaSpa~\cite{xia2025training} uses a fixed sparsity target and evaluates all blocks at scheduled search steps, whereas GSRS progressively lowers Top-p thresholds and restricts subsequent searches to the early candidate set $\mathcal{S}_1$.
Once LiteAttention~\cite{liteattention} skips a block, it excludes that block from evaluation and computation for the remaining denoising steps.
GSRS re-evaluates all blocks within $\mathcal{S}_1$ at each search step, allowing previously omitted blocks within $\mathcal{S}_1$ to be selected again as attention changes.
This design preserves the ability to dynamically select blocks while reducing search overhead compared with evaluating all blocks at each search step.

\paragraph{Block importance estimation.}
GSRS computes block importance from block log-sum-exp statistics and the current row-wise LSE, without relying on cached LSE values.
AdaSpa uses cached LSE values to estimate block importance, while LiteAttention applies a local tile-skipping condition based on the running softmax statistics.
The distinction between cached and current-step normalization affects both cumulative probability thresholds and block rankings.

Let $u$ and $v$ denote a query block and a key block, respectively, $B=|u|$ the number of query rows, and $z_{ij}$ the pre-softmax attention score.
For a fixed query block, let $\mathcal{C}$ denote the evaluated key blocks and $\mathcal{K}_{\mathcal{C}}$ their key tokens.
The current row-wise normalization factor is $L_i=\log\sum_{j\in\mathcal{K}_{\mathcal{C}}}\exp(z_{ij})$.
The normalized attention weight assigned to block $v\in\mathcal{C}$ and its query-block average are
\begin{equation}
    S_{iv}=\sum_{j\in v}\exp(z_{ij}-L_i),
    \qquad W_{uv}=\frac{1}{B}\sum_{i\in u}S_{iv},
    \qquad \sum_{v\in\mathcal{C}}S_{iv}=1.
\end{equation}
At subsequent search steps, AdaSpa reuses row-wise LSE values cached from an earlier step to estimate current block importance.
Denoting these cached values by $L_i^{\mathrm{old}}$, the resulting per-row block scores are
\begin{equation}
    \widetilde{S}_{iv}=\sum_{j\in v}\exp(z_{ij}-L_i^{\mathrm{old}})
    =\alpha_i S_{iv},
    \qquad \alpha_i=\exp(L_i-L_i^{\mathrm{old}}).
\end{equation}
Consequently, $\sum_{v\in\mathcal{C}}\widetilde{S}_{iv}=\alpha_i$, which need not equal one.
If such uncorrected scores are used for Top-p selection, the cumulative threshold can be reached too early or too late.
AdaSpa aggregates these scores over query rows to obtain block importance.
Because $\alpha_i$ varies across rows, this aggregation introduces query-dependent reweighting:
\begin{equation}
    \widetilde{W}_{uv}=\frac{1}{B}\sum_{i\in u}\alpha_i S_{iv},
    \qquad
    \widetilde{W}_{uv}-\widetilde{W}_{uw}
    =\frac{1}{B}\sum_{i\in u}\alpha_i(S_{iv}-S_{iw}).
\end{equation}
For two candidate blocks $v$ and $w$, this difference can have a different sign from $W_{uv}-W_{uw}$, changing their ranking and potentially affecting Top-k selection as well.
Normalizing the aggregated scores by a single scalar does not undo this query-dependent reweighting.

GSRS directly computes block importance using block log-sum-exp statistics $G_{iv}$ and the current row-wise LSE $L_i$:
\begin{equation}
    W_{uv}=\frac{1}{B}\sum_{i\in u}\exp(G_{iv}-L_i).
\end{equation}
This yields normalized block importance within the search candidate set without requiring cached LSE values.
The equivalence to the block importance definition is derived in Appendix~\ref{sec:app_block_importance_proof}.

LiteAttention compares each block's row-wise maximum attention scores with the running row-wise maxima maintained during online softmax.
A block is marked for skipping when its local maximum falls below the running maximum by a specified margin for every query row.
This criterion uses intermediate softmax statistics and depends on block processing order, which may lead to less accurate block selection.

\paragraph{Experimental comparison.}
Table~\ref{tab:app_related_methods} compares GSRS with related sparse attention methods on HunyuanVideo.
To match AdaSpa without head-adaptive allocation, we replace Top-p selection in GSRS with fixed Top-k selection using a uniform 20\% keep ratio for every head.
This configuration removes differences in Top-p selection and head-adaptive sparsity allocation from the comparison.
We also implement an Ampere-compatible LiteAttention kernel in CUDA, preserving its original skipping criterion and cross-step mask update mechanism.

\begin{table}[htbp]
    \centering
    \caption{Comparison with related sparse attention methods on HunyuanVideo.
    GSRS (AdaSpa aligned) and AdaSpa (w/o Head Adaptive) both use a uniform 20\% keep ratio.}
    \label{tab:app_related_methods}
    \begin{tabular}{lccc}
        \toprule
        Method & PSNR$\uparrow$ & SSIM$\uparrow$ & Speedup$\uparrow$ \\
        \midrule
        MInference & 22.53 & 0.7435 & 1.27$\times$ \\
        XAttention & 29.08 & 0.9099 & 1.36$\times$ \\
        LiteAttention & 21.85 & 0.7036 & 1.33$\times$\\
        AdaSpa (w/o Head Adaptive) & 28.51 & 0.8825 & 1.76$\times$ \\
        AdaSpa & 29.07 & 0.8905 & 1.78$\times$ \\
        \rowcolor{blue!8}
        \textbf{GSRS (AdaSpa aligned)} & \textbf{29.66} & \textbf{0.9330} & \textbf{1.86$\times$} \\
        \bottomrule
    \end{tabular}
\end{table}

Under the matched keep ratio, GSRS improves PSNR and SSIM over AdaSpa without head-adaptive allocation while achieving a higher speedup.
These results show that the gains are not solely attributable to Top-p selection or head-adaptive sparsity allocation.
GSRS also outperforms our Ampere implementation of LiteAttention in PSNR, SSIM, and speedup.
These results support the effectiveness of combining Reduced Search with precise block importance evaluation to reduce search overhead while preserving accurate block selection.

\newpage
\section{More Details about Attention Entropy and Mean Attention Distance}
\label{sec:metric_details}
To characterize the attention dynamics without incurring the $O(L^2)$ memory overhead of materializing full attention maps, we fuse the computation of attention entropy $H(i)$ and mean attention distance $D(i)$ directly into our FlashAttention-based kernels.
This section provides the formal derivation of the efficient calculation used in our implementation.

\subsection{Efficient Derivation of Attention Entropy}
The attention entropy for a specific query $i$ is defined based on the softmax-normalized weights $w_{ij}$:
\begin{equation}
H(i) = -\sum_{j=1}^{n} w_{ij} \log_2 w_{ij}
\end{equation}
In a hardware-efficient FlashAttention implementation, the weight $w_{ij}$ is typically computed using the raw attention score $s_{ij}$ (e.g., the scaled dot-product $QK^T/\sqrt{d}$) and the Log-Sum-Exp constant $\text{LSE}_i$:
\begin{equation}
w_{ij} = 2^{s_{ij} - \text{LSE}_i}
\end{equation}
Substituting this into the entropy formula, we can transform it into a form that avoids explicit exponentiation of every element during the final reduction:
\begin{equation} 
\begin{aligned}
H(i) &= -\sum_{j=1}^{n} w_{ij} (s_{ij} - \text{LSE}_i) \\
\end{aligned}
\end{equation}
Given the normalization property $\sum_{j=1}^n w_{ij} = 1$, the expression simplifies to:
\begin{equation}
H(i) = \text{LSE}_i - \mathbb{E}[s_i]
\end{equation}
where $\mathbb{E}[s_i] = \sum_j w_{ij} s_{ij}$ represents the expected value of the raw attention scores for query $i$.
This formulation allows us to compute entropy by maintaining a single running scalar $\sum w_{ij} s_{ij}$ during the block-wise tiling process.

\subsection{Fused Accumulation Logic}

Consistent with the definition in the Section~\ref{sec:analysis}, the Mean Attention Distance $D(i)$ represents the expected relative distance between the query $i$ and its attended keys $j$:
\begin{equation}
D(i) = \sum_{j} w_{ij} \cdot |i - j|
\end{equation}
In our Triton-based \cite{tillet2019triton} implementation, we fuse the calculation of $H(i)$ and $D(i)$ into the attention kernel by maintaining two running accumulators, $acc\_s$ and $acc\_d$:
\begin{enumerate}
    \item \textbf{Accumulation}: For each block of keys, we compute $acc\_s \leftarrow \sum w_{ij} s_{ij}$ and $acc\_d \leftarrow \sum w_{ij} |i - j|$.
    \item \textbf{Online Rescaling}: Since the softmax local maximum $m_i$ and the LSE update during the kernel's execution, we rescale $acc\_s$ and $acc\_d$ at each step to ensure numerical consistency, identical to the standard output $O$ rescaling in FlashAttention.
    \item \textbf{Finalization}: After the final reduction, $H(i)$ is obtained by subtracting the accumulated $acc\_s$ from the final $\text{LSE}_i$.
\end{enumerate}

This approach ensures that capturing these diagnostics adds negligible computational overhead and zero additional memory footprint beyond the storage of the resulting metrics themselves.

\subsection{Attention Dynamics Across Prompts}
\label{sec:app_prompt_analysis}
\begin{figure}[H]
    \centering
    \includegraphics[width=0.8\textwidth]{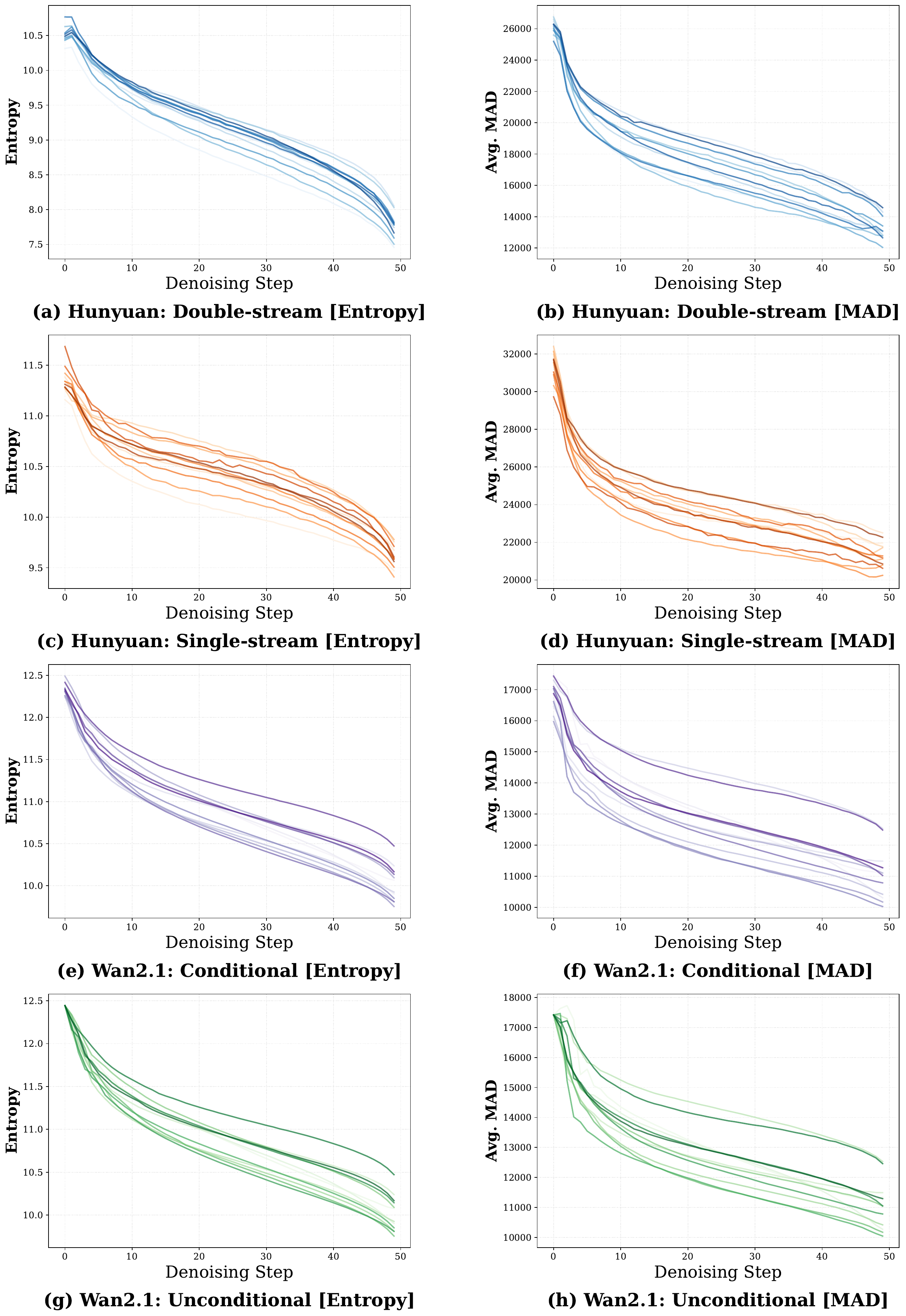}
    \caption{\textbf{Attention entropy and MAD profiles across diverse prompts for HunyuanVideo and Wan2.1.}
    Each line represents a different test prompt.
    Both metrics exhibit a consistent monotonic decrease across all prompts, confirming that the attention concentration trend is an intrinsic model property independent of input content.}
    \label{fig:detailed_mad_entropy}
\end{figure}

To verify that the attention trends reported in Section~\ref{sec:analysis} are not artifacts of specific inputs, we visualize attention entropy and mean attention distance across a diverse set of test prompts for both HunyuanVideo and Wan2.1 in Figure~\ref{fig:detailed_mad_entropy}.

Despite substantial differences in generated content, all prompts exhibit the same monotonic decrease in both metrics as denoising progresses.
The curves remain tightly bundled, with only minor variation in absolute values between prompts.
This confirms that the progressive concentration of attention is an intrinsic property of the model architecture rather than a content-dependent phenomenon.

\subsection{Attention Dynamics Across Layers}
\label{sec:app_layers_analysis}

\begin{figure}[H]
    \centering
    \includegraphics[width=0.7\textwidth]{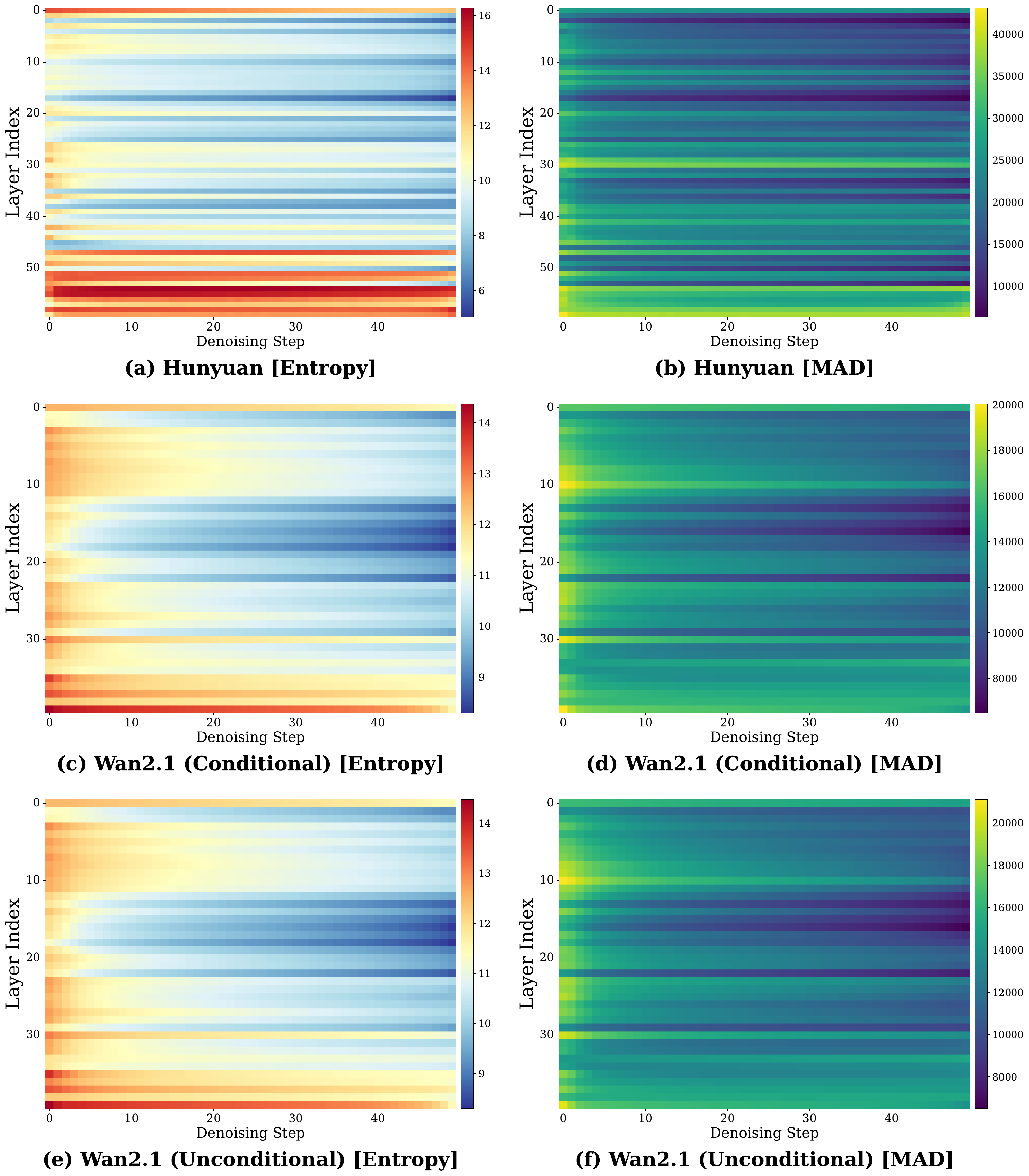}
    \caption{\textbf{Layer-wise attention entropy and MAD across denoising steps.}
    Both metrics decrease monotonically within each layer.
    Meanwhile, substantial variation exists across layers, confirming distinct attention concentration patterns at different depths.}

    \label{fig:layer_dynamics}
\end{figure}

Figure~\ref{fig:layer_dynamics} presents layer-wise attention entropy and mean attention distance (MAD) heatmaps for HunyuanVideo and Wan2.1.
Across all models, two consistent patterns emerge.
First, both metrics decrease monotonically along the denoising trajectory within each layer, confirming that the progressive concentration observed at the model level (Section~\ref{sec:analysis}) holds at the individual layer level as well.
Second, the variation across layers is substantial and persistent: certain layers maintain high entropy and long attention distance throughout denoising, while others exhibit consistently concentrated and localized attention.
Notably, for Wan2.1, the conditional and unconditional branches display similar layer-wise patterns, suggesting that this variation is an architectural property rather than a conditioning artifact.
These observations further support the need for layer-adaptive sparsity, as a uniform sparsity ratio would over-prune concentrated layers while under-pruning dispersed ones.

\subsection{Attention Dynamics Across Heads}
\label{sec:app_head_analysis}
Across heads, we visualize attention entropy and mean attention distance (MAD) for representative layers in HunyuanVideo and Wan2.1 (conditional and unconditional branches).
Each figure shows head-wise heatmaps over denoising steps.

Within each layer, different heads behave differently.
Some heads keep more dispersed attention, while others stay more concentrated.
Moreover, all heads show a similar decreasing trend as denoising progresses.
The conditional and unconditional branches of Wan2.1 also show very similar patterns, suggesting that these differences across heads are not caused by conditioning.
\begin{figure}[h]
    \centering
    \includegraphics[width=0.7\textwidth]{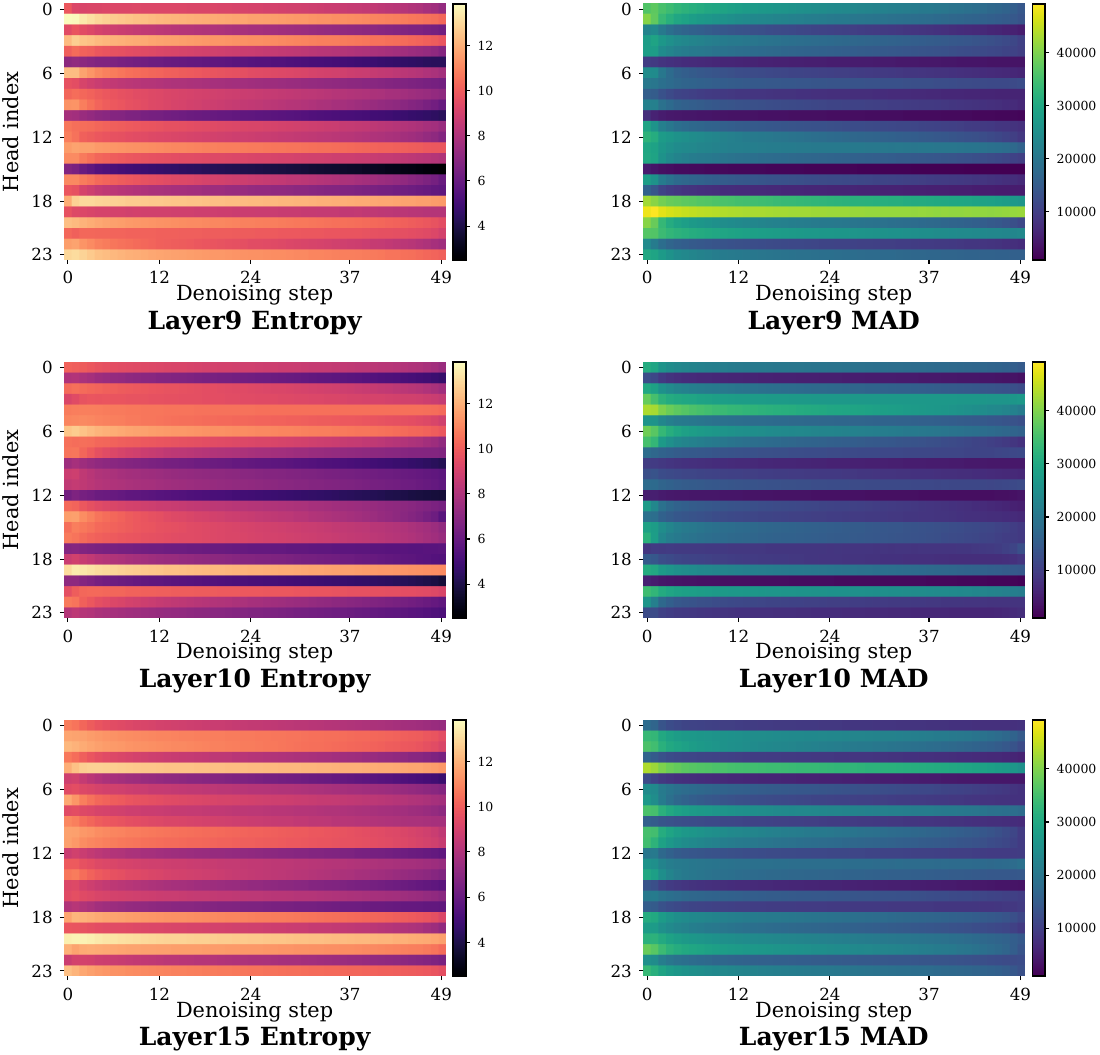}
    \caption{ \textbf{Head-level attention entropy and MAD in HunyuanVideo.}
    Each row corresponds to one layer, showing entropy (left) and MAD (right) for all heads.
    While different heads exhibit diverse behaviors, all follow a consistent decreasing trend across denoising steps.
    }
    \label{fig:head_hunyuan}
\end{figure}

\begin{figure}[h]
    \centering
    \includegraphics[width=0.7\textwidth]{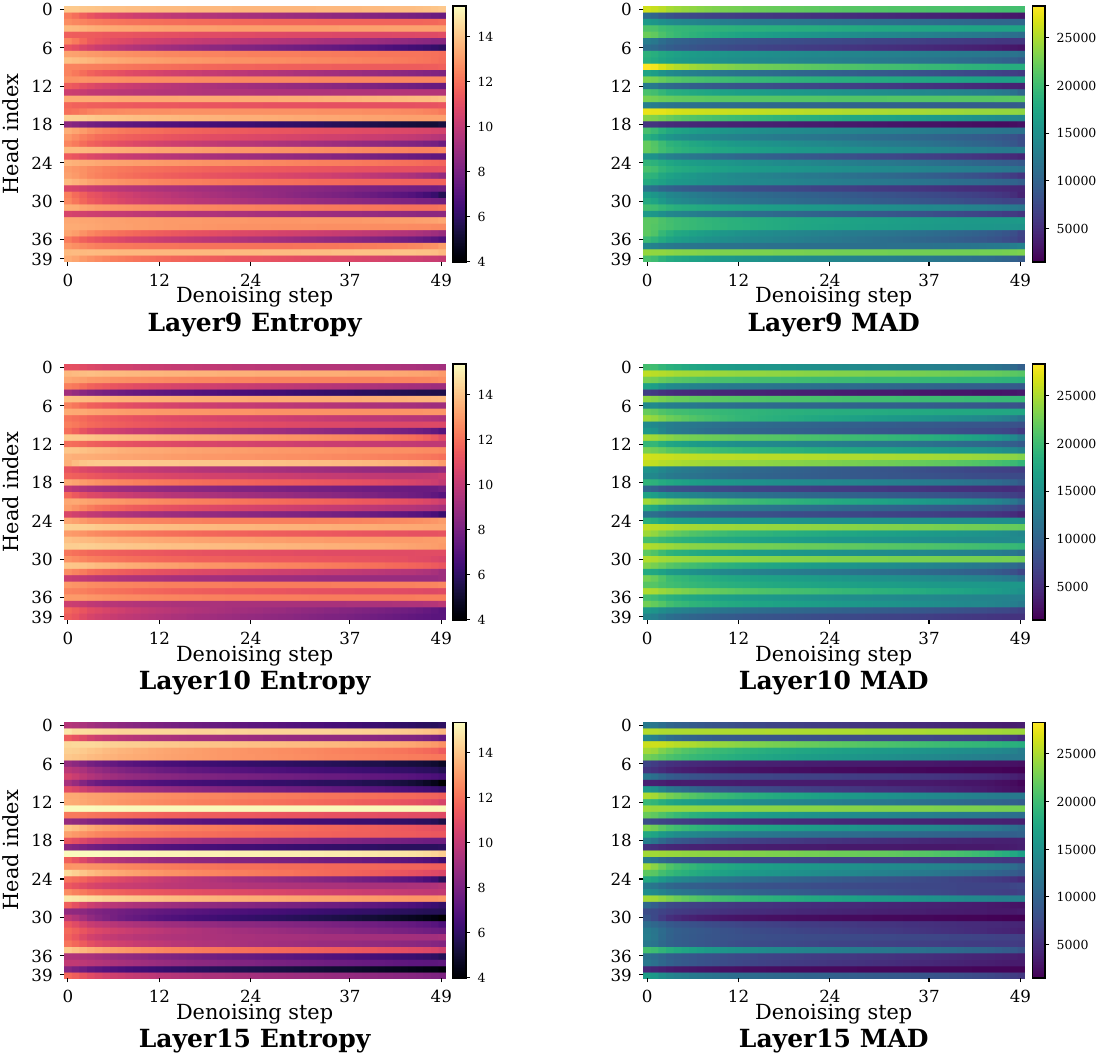}
    \caption{ \textbf{Head-level attention entropy and MAD in Wan2.1 (conditional branch).}
    Significant variation exists across heads within each layer, while the overall monotonic decrease over denoising is preserved.
    }
    \label{fig:head_wan_cond}
\end{figure}

\begin{figure}[t]
    \centering
    \includegraphics[width=0.7\textwidth]{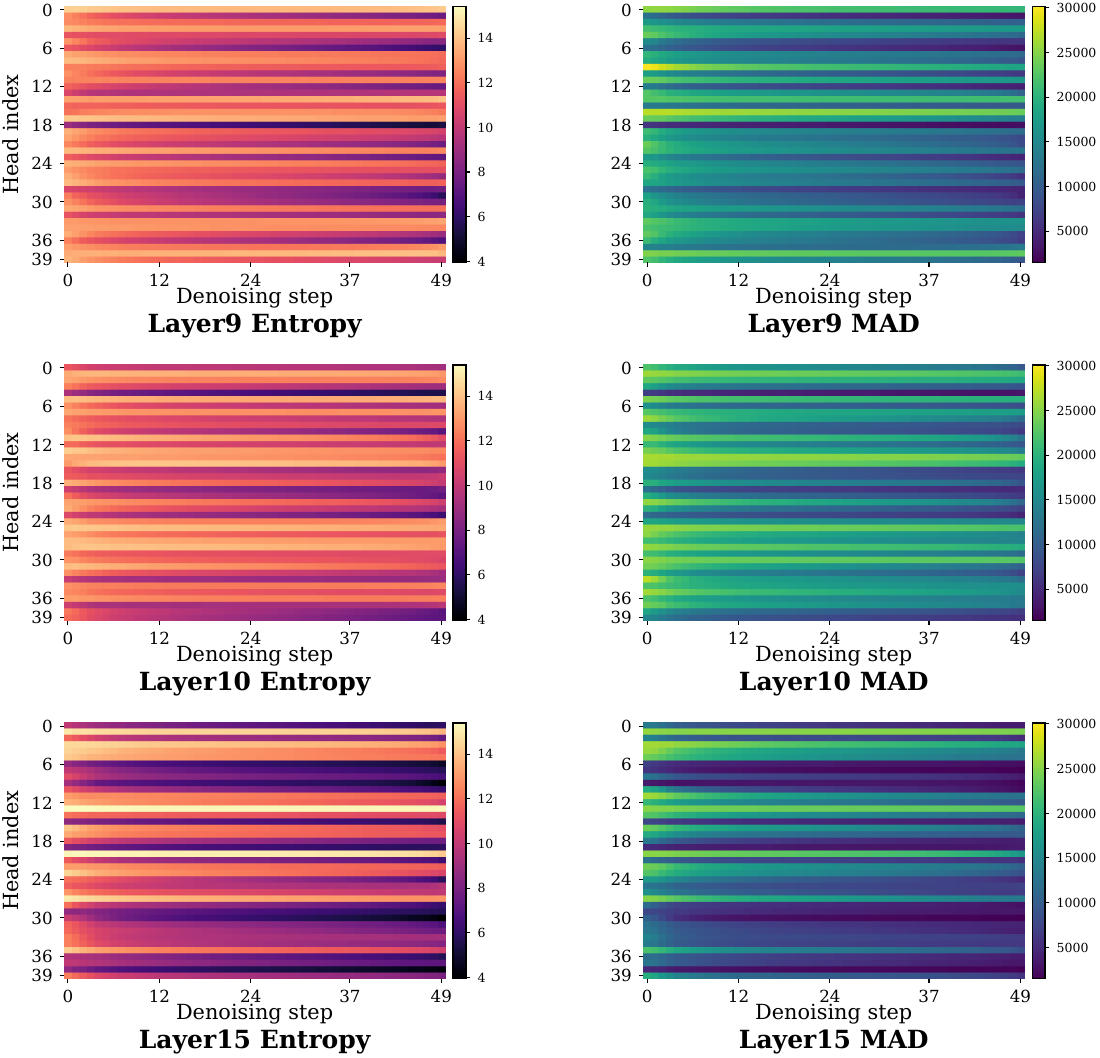}
    \caption{ \textbf{Head-level attention entropy and MAD in Wan2.1 (unconditional branch).}
    The patterns are highly consistent with the conditional branch, suggesting that head-wise variation is largely independent of conditioning.
    }
    \label{fig:head_wan_uncond}
\end{figure}

\clearpage

\section{Visualization of the generated videos}
\label{sec:visualization}
\begin{figure}[H]
    \centering
    \includegraphics[width=0.9\textwidth]{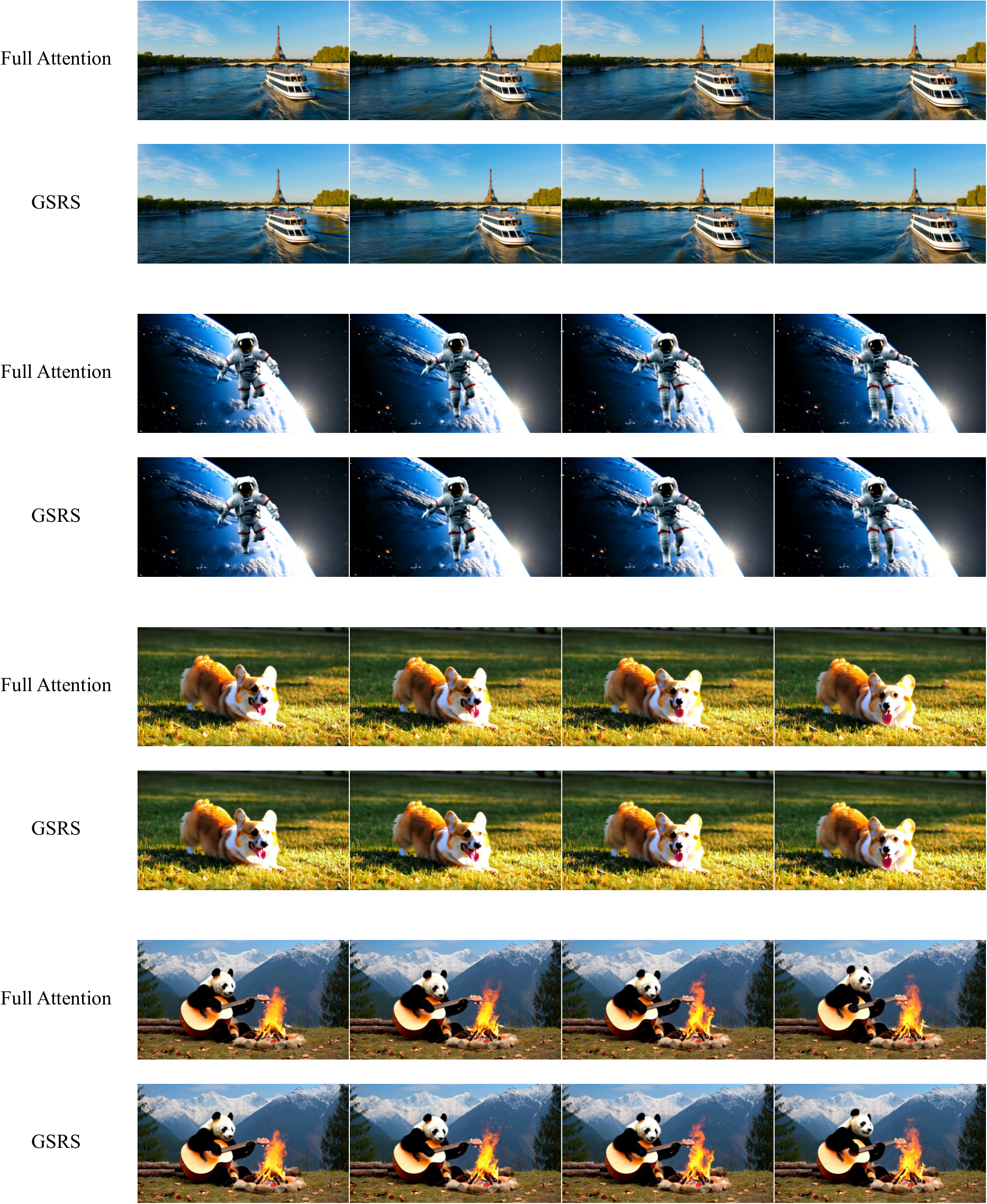}
    \caption{Comparion of Full Attention and GSRS on Wan 2.1 Text-to-Video generation.
    }
    \label{fig:wan_demo}
\end{figure}

\begin{figure}[H]
    \centering
    \includegraphics[width=0.9\textwidth]{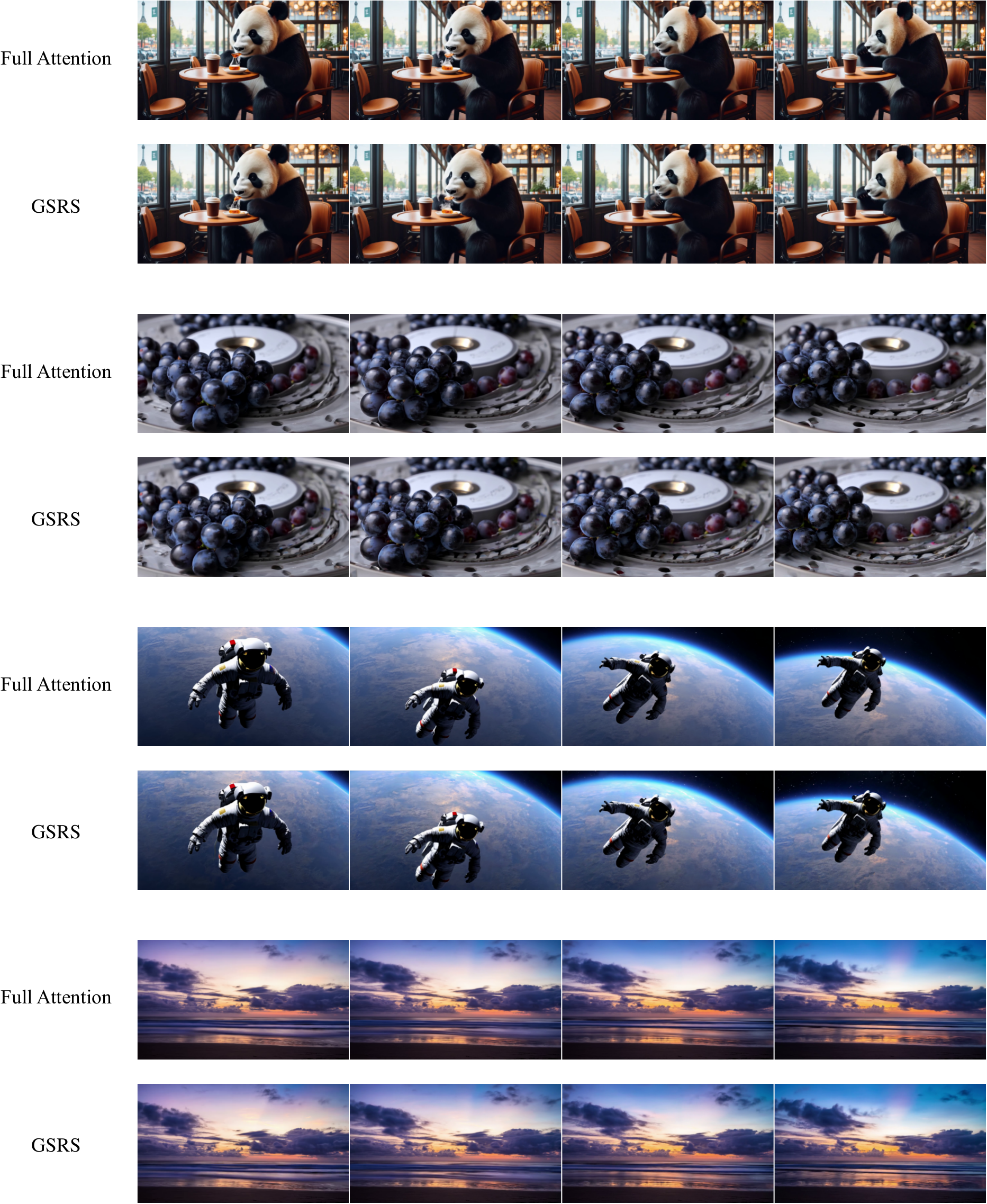}
    \caption{Comparion of Full Attention and GSRS on HunyuanVideo Text-to-Video generation.
    }
    \label{fig:hyvideo_demo}
\end{figure}

We provide visualization comparison between Full Attention and GSRS on Wan 2.1 and HunyuanVideo.
Results in Figure~\ref{fig:wan_demo} and Figure~\ref{fig:hyvideo_demo} demonstrate that GSRS can preserve high pixel-level fidelity, achieving similar generation quality compared with the full attention.
More video samples are provided in the supplementary materials.

\end{document}